%% file: arxiv.tex
\documentclass{article}

\PassOptionsToPackage{numbers}{natbib}

\usepackage[preprint]{neurips_2021}

\usepackage[utf8]{inputenc} 
\usepackage[T1]{fontenc}    
\usepackage{hyperref} 
\usepackage{url}            
\usepackage{booktabs}       
\usepackage{amsfonts}       
\usepackage{nicefrac}       
\usepackage{microtype}      
\usepackage{xcolor}         
\usepackage{graphicx}
\usepackage{amsmath}
\usepackage{amsthm}
\usepackage{tikz}
\usepackage{pgfplots}
\usetikzlibrary{arrows,decorations.pathmorphing,backgrounds,positioning,fit,
petri, decorations.pathreplacing,shadows,calc}
\usepackage{caption}
\usepackage{subcaption}

\definecolor{echoreg}{HTML}{2cb1e1}
\tikzset{%
  cascadedd/.style = {%
    general shadow = {%
      shadow scale = 1,
      shadow xshift = -3ex,
      shadow yshift = 3ex,
      draw=black,
      thick,
      fill = white},
    general shadow = {%
      shadow scale = 1,
      shadow xshift = -2.5ex,
      shadow yshift = 2.5ex,
      draw=black,
      thick,
      fill = white},
    general shadow = {%
      shadow scale = 1,
      shadow xshift = -2ex,
      shadow yshift = 2ex,
      draw=black,
      thick,
      fill = white},
    general shadow = {%
      shadow scale = 1,
      shadow xshift = -1.5ex,
      shadow yshift = 1.5ex,
      draw=black,
      thick,
      fill = white},
    general shadow = {%
      shadow scale = 1,
      shadow xshift = -1ex,
      shadow yshift = 1ex,
      draw=black,
      thick,
      fill = white},
    general shadow = {%
      shadow scale = 1,
      shadow xshift = -.5ex,
      shadow yshift = .5ex,
      draw=black,
      thick,
      fill = white},
    fill = white, 
    draw,
    thick}}
    \tikzset{%
    cascadedd_images/.style = {%
      general shadow = {%
        shadow scale = 1,
        shadow xshift = -3ex,
        shadow yshift = 3ex,
        draw=black,
        thick,
        fill = white},
      general shadow = {%
        shadow scale = 1,
        shadow xshift = -2.5ex,
        shadow yshift = 2.5ex,
        draw=black,
        thick,
        fill = white},
      general shadow = {%
        shadow scale = 1,
        shadow xshift = -2ex,
        shadow yshift = 2ex,
        draw=black,
        thick,
        fill = white},
      general shadow = {%
        shadow scale = 1,
        shadow xshift = -1.5ex,
        shadow yshift = 1.5ex,
        draw=black,
        thick,
        fill = white},
      general shadow = {%
        shadow scale = 1,
        shadow xshift = -1ex,
        shadow yshift = 1ex,
        draw=black,
        thick,
        fill = white},
      general shadow = {%
        minimum width=1cm,
        minimum height=1cm,
        shadow scale = 1,
        shadow xshift = -.5ex,
        shadow yshift = .5ex,
        draw=black,
        thick,
        fill = white},
      fill = white, 
      draw,
      thick}}
    
    \tikzset{%
              base/.style = {rectangle, rounded corners, draw=black,thick,
                             text centered,fill = echoreg!30, font=\sffamily},
    input/.style = {base,minimum width=2cm,align=center, minimum height=1cm,
    draw=black, text width=2cm},
    supportset/.style={cascadedd, align=center,rounded corners,fill = echoreg!30, 
    minimum width=2cm, minimum height=1cm, text width=2cm },
    arrow/.style = {very thick,-stealth},
    architecture/.style={opacity=0.8,align=center,rectangle, draw, 
    rounded corners, thick, fill=red!30, minimum width=2.5cm, minimum height=0.8cm},}
\title{Memory Wrap: a Data-Efficient and Interpretable Extension to Image Classification Models}

\author{%
  Biagio La Rosa\thanks{Contact Author} \\
  Sapienza University of Rome\\
  \texttt{larosa@diag.uniroma1.it} \\
  \And
  Roberto Capobianco \\
  Sapienza University of Rome\\ 
  Sony AI\\
  \texttt{capobianco@diag.uniroma1.it} \\
  \AND
   Daniele Nardi \\
   Sapienza University of Rome\\ 
  \texttt{nardi@diag.uniroma1.it} \\
}

\begin{document}

\maketitle

\begin{abstract}
  Due to their black-box and data-hungry nature, deep learning techniques are not yet widely adopted for real-world applications in critical domains, like healthcare and justice. This paper presents Memory Wrap, a plug-and-play extension to any image classification model. Memory Wrap improves both data-efficiency and model interpretability, adopting a content-attention mechanism between the input and some memories of past training samples. We show that Memory Wrap outperforms standard classifiers when it learns from a limited set of data, and it reaches comparable performance when it learns from the full dataset. We discuss how its structure and content-attention mechanisms make predictions interpretable, compared to standard classifiers. To this end, we both show a method to build explanations by examples and counterfactuals, based on the memory content, and how to exploit them to get insights about its decision process. 
  We test our approach on image classification tasks using several architectures on three different datasets, namely CIFAR10, SVHN and CINIC10.
\end{abstract}

\input{sections/introduction}
\input{sections/background}
\input{sections/method}
\input{sections/resultsPerformance}
\input{sections/resultsExplanations}
\input{sections/conclusion}

\begin{ack}
  This material is based upon work supported by Google Cloud.
\end{ack}

\bibliographystyle{named} 
\bibliography{biblio}

\input{appendix}

\end{document}

%% file: sections/introduction.tex
\section{Introduction}
\label{sec:introduction}
In the last decade, Artificial Intelligence has seen an explosion of
applications thanks to advancements in deep learning techniques. 
Despite their success, these techniques suffer from some important
problems: they require a lot of data to work well, and they act as
black boxes, taking an input and returning an output without providing
any explanation about that decision. The lack of transparency limits
the adoption of deep learning in important domains like health-care
and justice, while the data requirement makes harder its
generalization on real-world tasks. Few-shot learning methods and
explainable artificial intelligence (XAI) approaches address these
problems. The former studies the data requirement, experimenting on a
type of machine learning problem where the model can only use a
limited number of samples; the latter studies the problem of
transparency, aiming at developing methods that can explain, at least
partially, the decision process of neural networks. While there is an
extensive literature on each topic, few works explore methods that can
be used both on low data regime and that can provide explanations
about their outputs.

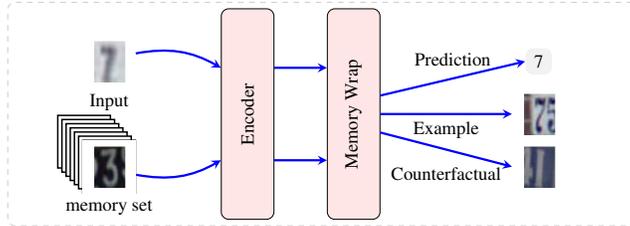
\begin{figure}[t!]
  \centering
  \scalebox{0.70}{
    \begin{tikzpicture} 
      \node(dummy1){};
      \node[inner sep=0pt, above of=dummy1,label=below:Input] (original) 
          {\includegraphics[width=1cm,height=1cm]{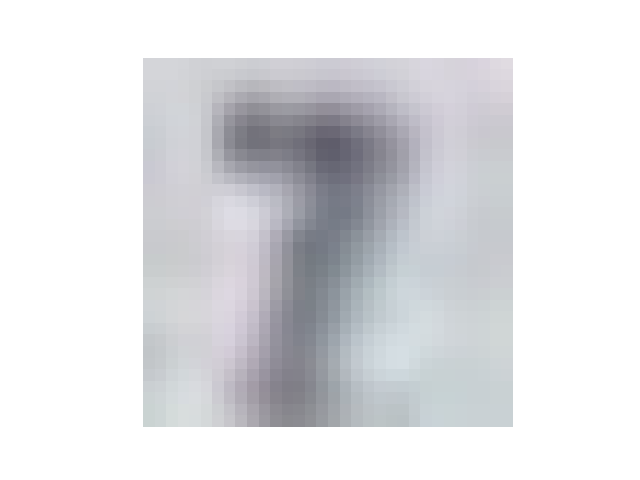}};
      \node[cascadedd_images,label=below:memory set,inner sep=0pt, below
      of=dummy1] (support)
      {\includegraphics[width=1cm,height=1cm]{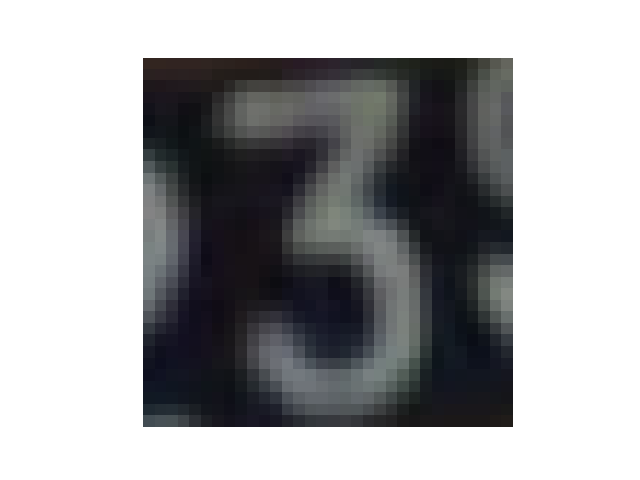}};
      \node[right=2cm of dummy1,rounded corners,rectangle,minimum width=1cm,
      minimum height=4cm,draw, fill=red!10](encoder){};
      \node[align=center,rotate=90] at (encoder.center) {Encoder};
      \node[right=1cm of encoder,rounded corners,rectangle,minimum width=1cm,
      minimum height=4cm,draw,fill=red!10](wrap){};
      \node[align=center,rotate=90] at (wrap.center) {Memory Wrap}; \node[inner
      sep=0pt, right=2.5cm of wrap] (example)
      {\includegraphics[width=1cm,height=1cm]{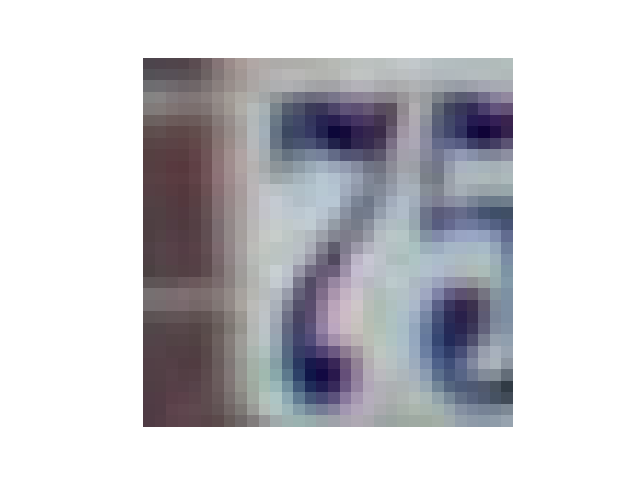}}; \node[above
      of=example, rectangle, rounded corners, minimum width=0.5cm, minimum
      height=0.5cm, fill=gray!10, opacity=80] (prediction) {7}; \node[inner
      sep=0pt, below of=example] (counter)
      {\includegraphics[width=1cm,height=1cm]{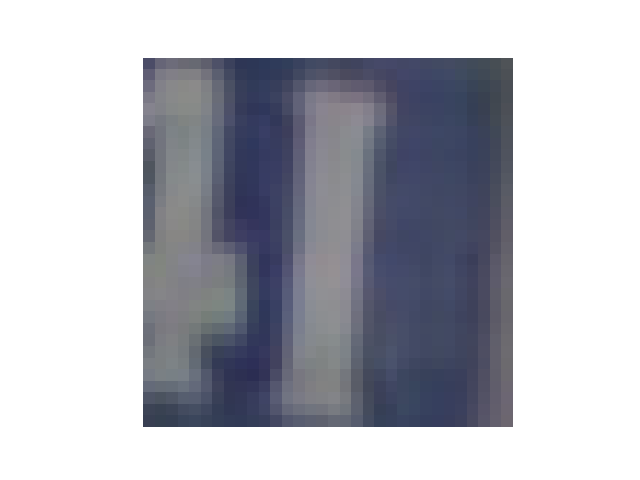}};
      \begin{scope}[on background layer]
        \node [draw=black!100, draw, rounded corners, dashed,minimum width=12cm,
        fill=white!20,opacity=0.2,fit={(counter)(encoder)(original)(support)}]
        {};
      \end{scope}
      \draw[arrow,bend left=20, blue]     (original) to
      ([yshift=25pt]encoder.west); \draw[arrow, bend right=20,blue]
      ([yshift=-25pt]support) to ([yshift=-25pt]encoder.west); \draw[arrow,
      blue]     ([yshift=25pt]encoder.east) -- ([yshift=25pt]wrap.west);
      \draw[arrow, blue]     ([yshift=-25pt]encoder.east) --
      ([yshift=-25pt]wrap.west); \draw[arrow, blue]     (wrap.east) --
      node[below, black] {Example} (example.west) ; \draw[arrow, blue]
      ([yshift=10pt]wrap.east) -- node[above=0.1cm,black] {Prediction}
      (prediction.west); \draw[arrow, blue]     ([yshift=-10pt]wrap.east) --
      node[left=0.2cm,below=0.2cm,black]{Counterfactual} (counter.west);
     
    \end{tikzpicture}
  }
    \label{fig:intro_image}
    \caption{Overview of Memory Wrap. The encoder takes as input an image and a
    memory set, containing random samples extracted from the training set. The
    encoder sends their latent representations to Memory Wrap, which outputs the
    prediction, an explanation by example, and a counterfactual, exploiting the
    sparse content attention between inputs encodings.}
\end{figure}

This paper makes a little step in both directions, proposing Memory
Wrap, an approach that makes image classification models more
data-efficient by providing, at the same time, a way to inspect their
decision process. In classical settings of supervised learning, models
use the training set only to adjust their weights, discarding it at
the end of the training process.  Instead, we hypothesize that, in a
low data regime, it is possible to strengthen the learning process by
re-using samples from the training set during inference.  Taking
inspiration from Memory Augmented Neural
Networks~\cite{Graves2014,Sukhbaatar2015}, the idea is to store a
bunch of past training samples (called \textit{memory set}) and
combine them with the current input through sparse attention
mechanisms to help the neural network decision process. Since the
network actively uses these samples during inference, we propose a
method based on inspection of sparse content attention weights to
extract insights and explanations about its predictions.

We test our approach on image classification tasks using
CIFAR10~\cite{Krizhevsky2009}, Street View House Number
(SVHN)~\cite{Netzer2011}, and CINIC10~\cite{Darlow2018} obtaining
promising results.  Our contribution can be summarized as follows:
\begin{itemize}
\item we present Memory Wrap, an extension for image classification
  models that uses a memory containing past training examples to
  enrich the input encoding;
\item we show it makes the original model more data-efficient,
  achieving higher accuracy on low data regimes;
\item we discuss methods to make their predictions more
  interpretable. In particular, we show that not only it is possible
  to extract the samples that actively contribute to the prediction,
  but we can also measure how much they contribute. Additionally, we
  show a method to retrieve similar examples from the memory that
  allow us to inspect which features are important for the current
  prediction, in the form of explanation by examples and
  counterfactuals.
\end{itemize}
The manuscript is organized as follows. Section 2 reviews existing
literature, focusing on works that use similar methods to us and
discuss the state-of-the-art in network explainability; Section 3
introduces our approach, while Section 4 presents some experiments and
their results. Finally, we discuss conclusions, limitations and future
directions.

%% file: sections/background.tex
\section{Background}
\label{sec:relatedwork}
\subsection{Memory Augmented Neural Networks}
\label{sec:relatedwork-mann}
Our work has been inspired by current advances in Memory Augmented
Neural Networks
(MANNs)~\cite{Graves2014,Graves2016,Kumar2016,Sukhbaatar2015}.  MANNs
use an external memory to store and retrieve data during input
processing. They can store past steps of a sequence, as in the case of
recurrent architectures for sequential tasks, or they can store
external knowledge in form of a knowledge base~\cite{Dinan2018}. Usually,
the network interacts with the memory through attention mechanisms,
and it can also learn how to write and read the memory during the
training process~\cite{Graves2014}. Differentiable Neural
Computers~\cite{Graves2016} and End-To-End Memory
Networks~\cite{Sukhbaatar2015} are popular examples of this class of
architectures. Researchers apply them to several problems like visual
question answering~\cite{Ma2018}, image classification~\cite{Cai2018},
and meta-learning~\cite{Santoro2016}, reaching great results.

Similarly to MANNs, Matching Networks~\cite{Vinyals2016} use a set of
\emph{never seen before} samples to boost the learning process of a
new class in one-shot classification tasks. Differently from us, their
architecture is standalone and it applies the product of attention
mechanisms on the labels of the sample set in order to compute the
final prediction.  Conversely, Prototypical Networks~\cite{Snell2017}
use samples of the training set to perform metric learning and to
return predictions based on the distance between prototypes in the
embedding space and the current input. Our approach relies on similar
ideas, but it uses a memory set that contains \emph{already
  seen and already learned} examples in conjunction with a sparse attention
mechanism. While we adopt a similarity measure to implement our
attention mechanism, we do not use prototypes or learned distances:
the network itself learns to choose which features should be retrieved
from each sample and which samples are important for a given
input. Moreover, our method differs from Prototype Networks because it
is model agnostic and can be potentially applied to any image
classification model without modifications.

\subsection{Explainable Artificial Intelligence}
\label{sec:relatedwork-xai}
\citet{Lipton2018} distinguishes between
\emph{transparent models}, where one can unfold the chain of reasoning
(e.g. decision trees), and \emph{post-hoc explanations}, that explain
predictions without looking inside the neural
network. The last category includes explanation by examples and
counterfactuals, which are the focus of our method.

\textbf{Explanations by examples} aim at extracting representative
instances from given data to show how the network
works~\cite{Belle2020}. Ideally, the instances should be similar to
the input and, in classification settings, predicted in the same
class. In this way, by comparing the input and the examples, a human
can extract both similarities between them and features that the
network uses to return answers.

\textbf{Counterfactuals} are specular to explanations by
examples: the instances, in this case, should be similar to the
current input but classified in another class. By comparing the input
to counterfactuals, it is possible to highlight differences and
to extract edits that should be applied to the current input to obtain
a different prediction. While for tabular data it is feasible to get
counterfactuals by changing features and at the same time to respect
domain constraints~\cite{Mahajan2019}, for images and natural language
processing the task is more challenging. This is due to the lack of
formal constraints and to the extremely large range of features to be
changed.

Recent research on explanation by examples and
counterfactuals adopts search methods~\cite{Wachter2017,Looveren2019},
which have high latency due to the large search space, and Generative
Adversarial Networks (GANs). For example, \citet{Liu2019}
use GANs to generate counterfactuals for images, but -- since they are
black-boxes themselves -- it is difficult to understand why a particular
counterfactual is a good candidate or not.

For small problems, techniques like KNN and SVM~\cite{Cortes1995} can
easily compute neighbors of the current input based on distance
measures, and use them as example-based explanations. Unfortunately,
for problems involving a large number of features and neural networks,
it becomes less trivial to find a correct distance metric that both
takes into account the different feature importance and that is
effectively linked to the neural network decision process. An attempt
in this direction is the twin-system proposed by \citet{Kenny2019},
which combines case-based reasoning systems (CBR) and neural
networks. The idea is to map the latent space or neural weights to
white-box case-based reasoners and extract from them explanations by
examples. With respect to these approaches, our method is intrinsic,
meaning that is embedded inside the architecture and, more
importantly, it is directly linked to the decision process, actively
contributing to it. Our method does not require external architectures
like GANs or CBR and it does not have any latency associated with its
use.

%% file: sections/method.tex
\section{Memory Wrap}
\label{sec:method}
This section describes the architecture of Memory Wrap and a methodology to
extract example-based explanations and counterfactuals for its
predictions.

\subsection{Architecture}
\label{sec:method-architecture}
Memory Wrap extends existing classifiers, specialized in a given task,
by replacing the last layer of the model. Specifically, it includes a
sparse content-attention mechanism and a multi-layer perceptron that
work together to exploit the combination of an input and a bunch of
training samples. In this way, the pre-existent model acts as an
\emph{encoder}, focused on extracting input features and mapping them
into a latent space. Memory Wrap stores previous examples
(\emph{memories}) that are then used at inference time. The only
requirement for the encoder is that its last layer -- before the Memory
Wrap -- outputs a vector containing a latent representation of the
input. Clearly, the structure of the encoder impacts on the
representation power, so we expect that a better encoder architecture
could improve further the performance of Memory Wrap.

More formally, let be $g(x)$ the whole model, $f(x)$ the encoder,
$x_i$ the current input, and $S_i=\{x^i_{m_1},x^i_{m_2},..,x^i_{m_n}\}$ a set of $n$
samples called \emph{memory set}, randomly extracted from the
training set during the current step $i$. First of all, the encoder
$f(x)$ encodes both the input and the memory set, projecting them in
the latent space and returning respectively:
\begin{align}
e_{x_i}&= f(x_i) & \mathbf{M}_{S_i} &= \{m^i_1, m^i_2,..,m^i_n\} = \{f(x^i_{m_1}),f(x^i_{m_2}),..,f(x^i_{m_n})\}.
\end{align}
Then, Memory Wrap computes the sparse content attention weights as the
sparsemax~\cite{Peters2019} of the similarity between the input
encoding and memory set encodings, thus attaching a content weight
$w_j$ to each encoded sample $m^i_j$.  We compute content attention
weights using the cosine similarity as in \citet{Graves2016}, replacing
the $softmax$ function with a $sparsemax$.\\
\begin{equation}
  \mathbf{w} = sparsemax(cosine[e_{x_i},\mathbf{M}_{S_i}]).
\end{equation} 
Since we are using the $sparsemax$ function, the memory vector only
includes information from few samples of the memory. In this way, each
sample contributes in a significant way, helping us to achieve output
explainability. Similarly to~\cite{Graves2016}, we compute the
\emph{memory vector} $\mathbf{v_{S_i}}$ as the weighted sum of memory
set encodings, where the weights are the content attention weights:
\begin{equation}
  \mathbf{v_{S_i}} = \mathbf{M}^T_{S_i}\mathbf{w}.
\end{equation}
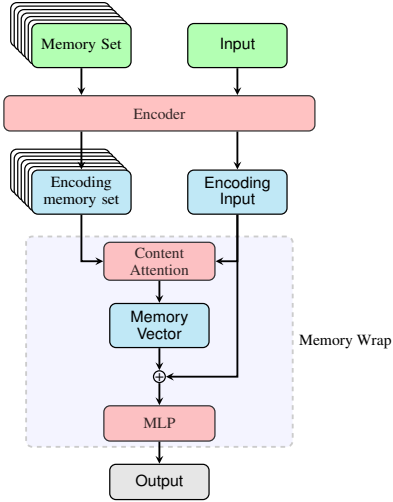
\begin{figure}[t!]
  \centering
  \resizebox{2.1in}{2.5in}{
\begin{tikzpicture}
    \node(dummy1){};
    \node [left=0.5cm of dummy1, supportset, fill=green!30] (ss) {Memory Set}; 
    \node [input, right= 0.5cm of dummy1, fill=green!30] (input) {Input};
    \node[architecture,below=1cm of dummy1,minimum width=7cm] (encoder) {Encoder};
    \node[below=1.25cm of encoder](dummy2){};
    \node [left=0.5cm of dummy2, supportset] (encoding_ss) {Encoding memory set}; 
    \node [input, right=0.5 of dummy2] (encoding_input) {Encoding Input};
    \node[below=0.5cm of dummy2](dummy3){};
    \node[architecture,below=1cm of dummy2,rounded corners, thick, fill=red!30](attention){Content\\Attention};
    \node[below=0.5cm of attention,input](sv){Memory Vector};
    \node[below=0.5cm of sv,circle,inner sep=0pt,minimum size=1pt, draw, thick](concat){+};
    \node[below=0.5cm of concat,architecture](mlp){MLP};
    \node[below=0.5cm of mlp,input,minimum height=0.8cm,fill=gray!20](output){Output};
    \begin{scope}[on background layer]
      \node [minimum width=6cm,label=right:{Memory Wrap}, draw=black!100, very thick,
      rounded corners, dashed, fill=blue!20,opacity=0.2,fit={(mlp) (concat) (attention)}] {};
    \end{scope}
    
    \draw[arrow]             (ss.south) -- (ss.south|-encoder.north);
    \draw[arrow]             (input.south) -| (input.south|-encoder.north);
    \draw[arrow]             (ss.south|-encoder.south) -- (encoding_ss.north);
    \draw[arrow]             (input.south|-encoder.south) -- (encoding_input.north);
    \draw[arrow]             (encoding_input.south) |- (attention.east);
    \draw[arrow]             (encoding_ss.south) |- (attention.west);
    \draw[arrow]             (attention.south) -| (sv);
    \draw[arrow]             (sv.south) -- (concat.north);
    \draw[arrow]             (encoding_input.south) |- (concat);
    \draw[arrow]             (concat) -- (mlp.north);
    \draw[arrow]             (mlp.south) -- (output.north);
 \end{tikzpicture}
  }
 \caption{Sketch of the system architecture. The system encodes the input and a
 bunch of training samples using a chosen neural network.
 Then, it generates a memory vector as a weighted sum of the memory set based on the sparse content attention weights between the encodings. Finally, the
 last layer predicts the input class, taking as input the concatenation of the memory vector and the input encoding.}
  \label{fig:teaser}
\end{figure}
Finally, the last layer $l_f$ takes the concatenation of the memory
vector and the encoded input, and returns the final output 
\begin{equation}
  o_i = g(x_i) =
l_f([e_(x_i),\mathbf{v_{S_i}}]).
\end{equation} 
The role of the memory vector is to enrich the input
encoding
with \emph{additional} features extracted from similar samples,
possibly missing on the current input. On average, considering the
whole memory set and thanks to the cosine similarity, strong features
of the target class will be more represented than features of other
classes, helping the network in the decision process. In our case, we
use a multi-layer perceptron with only one hidden layer as a final
layer, but other choices are possible (App.~\ref{appx:finallayer}).



\subsection{Getting explanations}
\label{sec:method-explanation}
We aim at two types of explanations: explanation by examples and
counterfactuals. The idea is to exploit the memory vector and content
attention weights to extract explanations about model outputs, in a
similar way to \citet{LaRosa2020}. To understand how, let's consider
the current input $x_i$, the current prediction $g(x_i)$, and the
encoding matrix $M_{S_i}$ of the memory set, where each
$m^i_j \in M_{S_i}$ is associated with a weight $w_j$. We can split
the matrix $M_{S_i}$ into three disjoint sets
$ M_{S_i} = M_e \cup M_c \cup M_z$, where
$M_e = \{f(x^i_{m_j}) \mid g(x_i) = g(x^i_{m_j})\}$ contains encodings
of samples predicted in the same class $g(x_i)$ by the network and
associated with a weight $w_j > 0$,
$M_c = \{f(x^i_{m_j}) \mid g(x_i) \neq g(x^i_{m_j})\}$ contains
encodings of samples predicted in a different class and associated
with a weight $w_j > 0$, and $M_z$ contains encodings of samples
associated with a weight $w_j = 0$. Note that this last set does not
contribute at all to the decision process and it cannot be considered
for explainability purposes.  Conversely, since $M_e$ and $M_c$ have
positive weights, they can be used to extract explanation by examples
and counterfactuals.


Let's consider, for each set, the sample $x^i_{m_j}$ associated with
the highest weight. A high weight of $w_j$ means that the encoding of
the input $x_i$ and the encoding of the sample $x^i_{m_j}$ are
similar. If $x^i_{m_j} \in M_e$, then it can be considered as a good
candidate for an explanation by example, being an instance similar to
the input and predicted in the same class, as defined in
Sect.~\ref{sec:relatedwork-xai}. Instead, if $x^i_{m_j} \in M_c$, then
it is considered as a counterfactual, being similar to the input but
predicted in a different class.  Finally, consider the sample
$x^i_{m_k}$ associated with the highest weight in the whole set
$M_{S_i}$. Because it is the highest, it will be heavily represented
in the memory vector that will \textbf{actively} contribute to the
inference, being used as input for the last layer. This means that
common features between the input and the sample $x^i_{m_k}$ are
highly represented and so they constitute a good
explanation. Moreover, if $x^i_{m_k}$ was a counterfactual, because it
is partially included in the memory vector, it is likely that it will
be the second or third predicted class, giving also information about
``doubts'' of the neural network.

%% file: sections/resultsPerformance.tex
\section{Results}
\label{sec:results}
This section first describes the experimental setups, and then it presents and
analyzes the obtained results for both performance and explanations.
\subsection{Setup}
\label{sec:results-setup}
We test our approach on image classification tasks using the Street
View House Number (SVHN)~\cite{Netzer2011}, CINIC10~\cite{Darlow2018}
and CIFAR10~\cite{Krizhevsky2009} datasets. For the encoder $f(x)$, we
run our tests using ResNet18~\cite{He2016}, EfficientNet
B0~\cite{Tan2019}, MobileNet-v2~\cite{Howard2017}, and other
architectures whose results are reported in
App.~\ref{appx:architectures}. We randomly split the training set to
extract smaller sets in the range \{1000,2000,5000\}, thus simulating
a low regime data setting, and then train each model using these sets
and the whole dataset. At each training step, we randomly extract 100
samples from the training set and we use them as memory set ---
$\sim$10 samples for each class (see App.~\ref{appx:memorysize} and
App.~\ref{appx:cost} for further details about this choice). We run 15
experiments for each configuration, fixing the seeds for each run and
therefore training each model under identical conditions. We report
the mean accuracy and the standard deviation over the 15 runs for each
model and dataset. For further details about the training setup please
consult App.~\ref{appx:trainingdetails}.
\subsubsection{Baselines}
\label{sec:results-baselines}
\paragraph{Standard.} This baseline is obtained with the classifiers $f(x)$ without
any modification and trained in the same manner of Memory Wrap
(i.e. same settings and seeds).
\paragraph{Only Memory.} This baseline uses only the memory vector
as input of the multi-layer perceptron, removing the
concatenation with the encoded input. Therefore, the output
is given by $o_i = g(x_i) = l_f(\mathbf{v_{S_i}})$. In this case, the
input is used only to compute the content weights, which are then used 
to build the memory vector, and
the network learns to predict the correct answer based on
them. Because of the randomness of the memory set and the absence of
the encoded input image as input of the last layer, the network is
encouraged to learn more general patterns and not to exploit specific
features of the given image.

\subsection{Performance}
\label{sec:results-performance}
\begin{table*}[bh!]
    \centering
    \caption{Avg. accuracy and standard deviation over 15 runs of the baselines and Memory Wrap, when the training dataset is a subset of SVHN. For each configuration, we highlight in bold the best result and results that are within its margin.}
    \label{table:reduced-SVHN}
    \begin{tabular}{@{}rrrrrr@{}}
      \toprule
      \multicolumn{5}{c}{\textbf{Reduced SVHN Avg. Accuracy\%}}\\
      \midrule
       &&\multicolumn{3}{c}{\textbf{Samples}}\\
      Model&Variant&1000&2000&5000\\
      \midrule
      \quad EfficientNetB0  & Standard  & 57.70 $\pm$ 7.89  & 72.59 $\pm$ 4.00  & 81.89 $\pm$ 3.37 \\
      \quad &Only Memory     & 58.86 $\pm$ 3.30  & 75.79 $\pm$ 1.68  & \textbf{85.30 $\pm$ 0.52} \\
      \quad &Memory Wrap     & \textbf{66.78 $\pm$ 1.27}  & \textbf{77.37 $\pm$ 1.25}  & \textbf{85.55 $\pm$ 0.59} \\
      \midrule
      \quad MobileNet-v2  &  Standard   & 42.71 $\pm$ 10.31  & 70.87 $\pm$ 4.20  & 85.52 $\pm$ 1.16 \\
      \quad &Only Memory      & 60.60 $\pm$ 3.14  & \textbf{80.80 $\pm$ 2.05}  & \textbf{88.77 $\pm$ 0.42} \\
      \quad & Memory Wrap       & \textbf{66.93 $\pm$ 3.15}  & \textbf{81.44 $\pm$ 0.76}  & \textbf{88.68 $\pm$ 0.46} \\
      \midrule
      \quad ResNet18  & Standard  & 20.63 $\pm$ 2.85  & 31.84 $\pm$ 18.38  &  79.03 $\pm$ 12.89 \\
      \quad & Only Memory      & 35.57 $\pm$ 6.48  & 68.87 $\pm$ 8.70  &  \textbf{87.63 $\pm$ 0.42} \\
      \quad & Memory Wrap     & \textbf{45.31 $\pm$ 8.19}  & \textbf{77.26 $\pm$ 3.38} & \textbf{87.74 $\pm$ 0.35} \\
      \bottomrule
    \end{tabular}
  \end{table*}
  \begin{table*}[bh!]
      \centering
      \caption{Avg. accuracy and standard deviation over 15 runs of the baselines and Memory Wrap, when the training dataset is a subset of CIFAR10. For each configuration, we highlight in bold the best result and results that are within its margin.}
      \label{table:reduced-CIFAR10}
      \begin{tabular}{@{}rrrrrr@{}}
        \toprule
        \multicolumn{5}{c}{\textbf{Reduced CIFAR10 Avg. Accuracy\%}}\\
        \midrule
         &&\multicolumn{3}{c}{\textbf{Samples}}\\
        Model&Variant&1000&2000&5000\\
        \midrule
        \quad EfficientNetB0  & Standard & 39.63 $\pm$ 2.16  & 47.25  $\pm$ 2.22  & 67.34 $\pm$ 2.37 \\
        \quad &Only Memory     & 40.60 $\pm$  2.04 & \textbf{52.87 $\pm$ 2.07}  & \textbf{70.82 $\pm$ 0.52} \\
        \quad &Memory Wrap    & \textbf{41.45 $\pm$ 0.79}  & \textbf{52.83 $\pm$ 1.41}  & \textbf{70.46 $\pm$ 0.78}\\
        \midrule
        \quad MobileNet-v2  &  Standard  & 38.57 $\pm$ 2.11  & 50.36 $\pm$  2.64 &  72.77 $\pm$ 2.21 \\
        \quad &Only Memory     & \textbf{43.15 $\pm$ 1.35}  & \textbf{57.43 $\pm$  1.45} & \textbf{75.56 $\pm$ 0.76} \\
        \quad & Memory Wrap      & \textbf{43.87 $\pm$ 1.40}  & \textbf{57.12 $\pm$ 1.36}  & \textbf{75.33 $\pm$ 0.62} \\
        \midrule
        \quad ResNet18  & Standard & \textbf{40.03 $\pm$ 1.36}  & 48.86 $\pm$ 1.57  &   65.95  $\pm$ 1.77 \\
        \quad & Only Memory     & \textbf{40.35 $\pm$ 0.89}  & \textbf{51.11 $\pm$ 1.22}  &   \textbf{70.28  $\pm$ 0.80}  \\
        \quad & Memory Wrap    & \textbf{40.91 $\pm$ 1.25}  &  \textbf{51.11 $\pm$ 1.13}  &  \textbf{69.87 $\pm$ 0.72} \\
        \bottomrule
      \end{tabular}
    \end{table*}
    \begin{table*}[bh!]
        \centering
        \caption{Avg. accuracy and standard deviation over 15 runs on SVHN
        dataset of the baselines and Memory Wrap. The
        comparison highlights the difference in performance when using a subset of
        the dataset in the range \{1k,2k,5k\} as training set. For each configuration, we highlight in bold the best result and results that are within its margin.}
        \label{table:reduced-CINIC10}
        \begin{tabular}[t]{@{}rrrrrr@{}}
          \toprule
          \multicolumn{5}{c}{\textbf{Reduced CINIC10 Avg. Accuracy\%}}\\
          \midrule
           &&\multicolumn{3}{c}{\textbf{Samples}}\\
          Model&Variant&1000&2000&5000\\
          \midrule
          \quad EfficientNetB0  & Standard  & 29.50 $\pm$ 1.18 & 33.56 $\pm$ 1.26  & 45.98 $\pm$ 1.34 \\
          \quad &Only Memory    & \textbf{30.46 $\pm$ 1.17} & \textbf{36.17 $\pm$  1.54} & 44.97 $\pm$ 0.95 \\
          \quad &Memory Wrap     & \textbf{30.45 $\pm$ 0.64} & \textbf{36.65 $\pm$ 1.03}	  & \textbf{47.06 $\pm$ 0.91}\\
          \midrule
          \quad MobileNet-v2  & Standard  & 29.61 $\pm$ 0.89 & 36.40 $\pm$ 1.58  & 50.41 $\pm$ 1.01 \\
          \quad & Only Memory      & \textbf{32.46 $\pm$ 1.07} & \textbf{39.91 $\pm$ 0.82}  & \textbf{52.51 $\pm$ 0.77}\\
          \quad & Memory Wrap     & \textbf{32.34 $\pm$ 0.95} & \textbf{39.48 $\pm$  1.16}  & \textbf{52.18 $\pm$ 0.66}\\
          \midrule
          \quad ResNet18  &  Standard   & 31.18 $\pm$ 1.21 & \textbf{37.67 $\pm$ 0.98}  & 45.39 $\pm$ 1.07 \\
          \quad & Only Memory     & 30.79 $\pm$ 0.83 & 37.30 $\pm$ 0.57  &  \textbf{46.66 $\pm$ 0.81}\\
          \quad & Memory Wrap       & \textbf{32.15 $\pm$ 0.68} & \textbf{38.51 $\pm$ 0.96}  & \textbf{46.39 $\pm$ 0.67} \\
          \bottomrule
        \end{tabular}
    \end{table*}
    \begin{table}[bh!]
      \setlength{\tabcolsep}{4pt}
        \centering
        \caption{Avg. accuracy and standard deviation over 15 runs of the baselines and Memory Wrap, when the training datasets are the whole SVHN and CIFAR10 datasets. For each configuration, we highlight in bold the best result and results that are within its margin.}
        \label{table:complete}
        \begin{tabular}{@{}rrrrr@{}}
          \toprule
          &\multicolumn{3}{c}{\textbf{Full Datasets Avg. Accuracy \%}}\\
          \midrule
          Model&Variant&SVHN&CIFAR10&CINIC10\\
          \midrule
          \quad EfficientNetB0& Standard &  94.39 $\pm$ 0.24  & \textbf{88.13 $\pm$ 0.38} & \textbf{77.31 $\pm$ 0.35}\\
          \quad & Only Memory  &  \textbf{94.63 $\pm$ 0.33}  & 86.48 $\pm$ 0.29 & 76.19 $\pm$ 0.25\\
          \quad & Memory Wrap  & \textbf{94.67 $\pm$ 0.16} & \textbf{88.05 $\pm$ 0.20} & \textbf{77.34 $\pm$ 0.27}\\
          \midrule
          \quad MobileNet-v2&Standard & \textbf{95.95 $\pm$ 0.09}  & \textbf{88.78 $\pm$ 0.41} & \textbf{78.97 $\pm$ 0.31} \\
          \quad & Only Memory &  95.59 $\pm$ 0.11  & 86.37 $\pm$ 0.21  & 74.60 $\pm$ 0.13\\
          \quad & Memory Wrap  &  95.63 $\pm$ 0.08  & \textbf{88.49 $\pm$ 0.32}  & \textbf{79.05 $\pm$ 0.15}\\
          \midrule
          \quad ResNet18 & Standard & \textbf{95.79 $\pm$ 0.18} & \textbf{91.94 $\pm$ 0.19} & \textbf{82.05 $\pm$ 0.25} \\
          \quad & Only Memory &  \textbf{95.82 $\pm$ 0.10}  & 91.36 $\pm$ 0.24  & 81.65 $\pm$ 0.19\\
          \quad & Memory Wrap  &  95.58 $\pm$ 0.06  & 91.49 $\pm$ 0.17  & \textbf{82.04 $\pm$ 0.16}\\
          \bottomrule
        \end{tabular}
      \end{table}

      In low data regimes, our method outperforms the standard models in all the
      datasets, sometimes with a large margin (Table~\ref{table:reduced-SVHN},
      Table~\ref{table:reduced-CINIC10}, and Table~\ref{table:reduced-CIFAR10}).
      First, we can observe that the amount of gain in performance depends on
      the used encoder: MobileNet shows the largest gap in all the datasets,
      while ResNet shows the smallest one, representing a challenging model for
      Memory Wrap. Secondly, it depends on the dataset, being the gains in each
      SVHN configuration always greater than the ones in CIFAR10 and CINIC10.
      Regarding the baseline that uses only the memory, it outperforms the
      standard baseline too, reaching nearly the same performance of Memory Wrap
      in most of the settings. However, its performance appears less stable
      across configurations, being lower than Memory Wrap in some SVHN and
      CINIC10 settings (Table~\ref{table:reduced-SVHN} and Table ~\ref{table:reduced-CINIC10}) and lower than standard
      models in some full dataset scenarios and in some configurations of CINIC10. These considerations are confirmed also on other architectures
      reported in App.~\ref{appx:architectures}. We hypothesize that the additional
      information captured by the input encoding allow the model to exploit
      additional shortcuts and to reach the best performance.

      Note that it is possible to increase the gap by adding more
      samples in the memory, at the cost of an increased training and
      inference time (App.~\ref{appx:memorysize}). Moreover, while in
      low data regimes standard neural networks performances show high
      variance, Memory Wrap seems to be a lot more stable with a lower
      standard deviation.
 
      When Memory Wrap learns from the full dataset (Table
      \ref{table:complete}), it reaches comparable performance most of
      the time. Hence, our approach is useful also when used
      with the full dataset, thanks to the additional interpretability
      opportunity provided by its structure
      (Section~\ref{sec:method-explanation}).

%% file: sections/resultsExplanations.tex
\subsection{Explanations}
\label{sec:results-explanations}

\begin{figure}[t!]
  \begin{subfigure}{.32\columnwidth}
    \centering
    \includegraphics[width=.9\columnwidth]{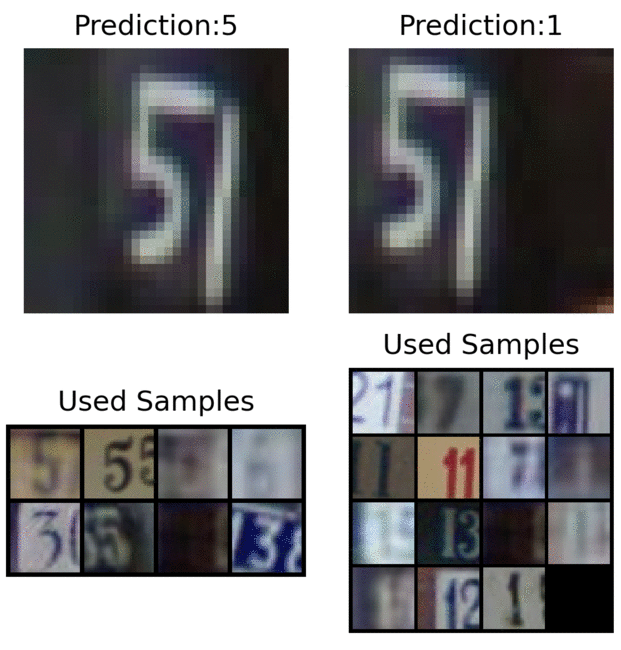}
    \caption{\label{fig:mem_svhn}}

  \end{subfigure}%
  \begin{subfigure}{.32\columnwidth}
    \centering
    \includegraphics[width=.9\columnwidth]{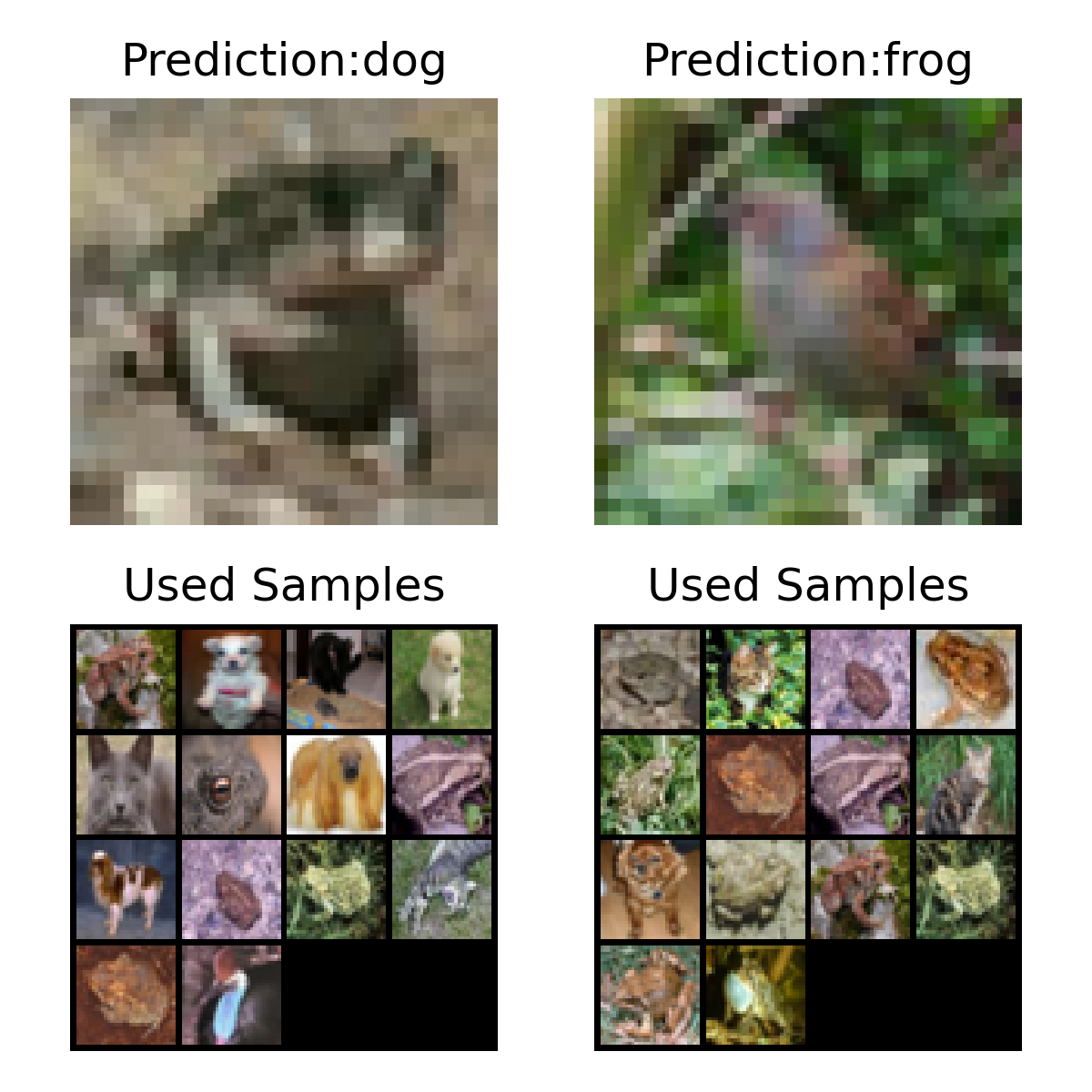}
    \caption{\label{fig:mem_cifar}}

  \end{subfigure}
  \begin{subfigure}{.32\columnwidth}
    \centering
    \includegraphics[width=.9\columnwidth]{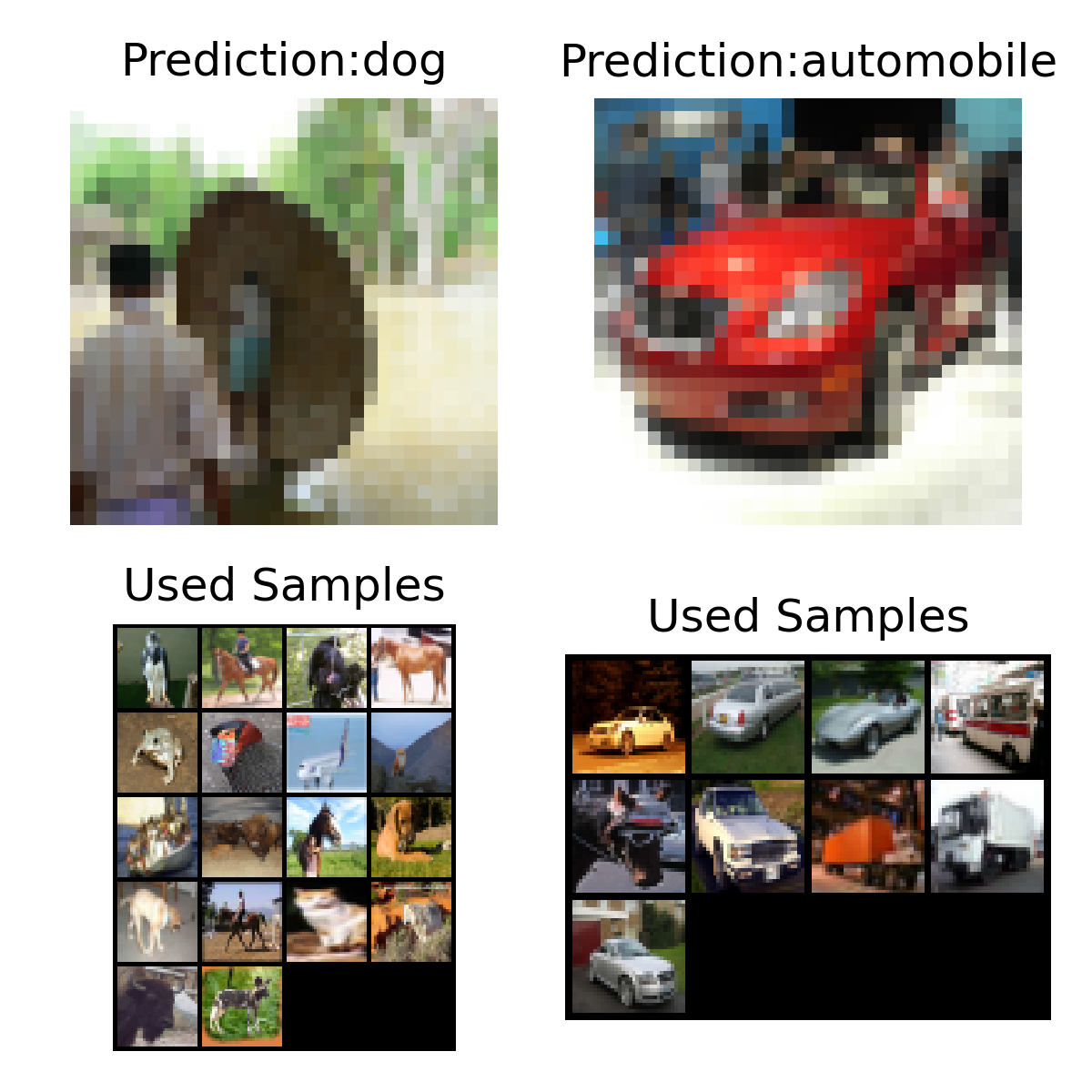}
    \caption{\label{fig:mem_cinic}}

  \end{subfigure}
  \caption{Inputs (first rows), their associated predictions and an overview of
  the samples in the memory set that have an active influence on the decision process
  -- i.e. the samples on which the memory vector is built -- (second row).}
  \label{fig:mem_example}
\end{figure}
\begin{table}[bh!]
    \centering
    \caption{Mean Explanation accuracy and standard deviation over 15 runs of the
    sample in the memory set with the highest sparse content attention weight.}
    \label{table:expaccuracy}
    \begin{tabular}{@{}rrrr@{}}
      \toprule
      &\multicolumn{2}{c}{\textbf{Explanation Accuracy \%}}\\
      \midrule
      Dataset&Samples&Only Memory&Memory Wrap\\
      \midrule
      \quad SVHN&1000 & 81.46 $\pm$ 1.05 & 84.26 $\pm$ 1.29\\
      \quad &2000 &  89.53 $\pm$ 1.05 & 90.76 $\pm$ 0.50\\
      \quad &5000 & 93.79 $\pm$ 0.23 & 94.56 $\pm$ 0.20\\
      \midrule
      \quad CIFAR10&1000 & 77.81 $\pm$ 0.93 & 78.03 $\pm$ 0.81\\
      \quad &2000 & 82.39 $\pm$ 0.65 & 82.01 $\pm$ 0.49\\
      \quad &5000 & 89.17 $\pm$ 0.34 & 88.49 $\pm$ 0.26\\
      \midrule
      \quad CINIC10&1000 & 75.99 $\pm$ 0.80 &  76.30 $\pm$ 0.75\\
      \quad &2000& 78.38 $\pm$ 0.46& 78.43 $\pm$ 0.55\\
      \quad &5000 & 80.90 $\pm$ 0.33& 74.47 $\pm$ 0.62\\
      \bottomrule

    \end{tabular}
\end{table}

\begin{table}[bh!]
    \centering
    \caption{Accuracy reached by the model on images where the sample with the
    highest weight in memory set is a counterfactual. The accuracy is computed as
    the mean over 15 runs using as encoder MobileNet-v2.}
    \label{table:counteraccuracy}
    \begin{tabular}{@{}rrrr@{}}
      \toprule
      &\multicolumn{2}{c}{\textbf{Accuracy \%}}\\
      \midrule
      Dataset&Samples&Only Memory&Memory Wrap\\
      \midrule
      \quad SVHN&1000 & 35.45 $\pm$ 2.47 & 39.90 $\pm$ 1.84\\
      \quad &2000 &  43.79 $\pm$ 1.20 & 45.60 $\pm$ 1.08\\
      \quad &5000 & 48.02 $\pm$ 1.02 & 49.02 $\pm$ 1.39\\
      \midrule
      \quad CIFAR10&1000 & 30.56 $\pm$ 1.08 & 31.78 $\pm$ 0.82\\
      \quad &2000 & 36.93 $\pm$ 0.83 & 37.15 $\pm$ 0.70\\
      \quad &5000 & 44.57 $\pm$ 1.19 & 45.80 $\pm$ 0.77\\
      \midrule
      \quad CINIC10&1000 & 25.20 $\pm$ 0.57 & 25.36 $\pm$ 0.62\\
      \quad &2000& 28.77 $\pm$ 0.42& 28.70 $\pm$ 0.59\\
      \quad &5000 & 34.18 $\pm$ 0.56 & 35.22 $\pm$ 0.46\\
      \bottomrule
      \end{tabular}
\end{table}

From now on, we will consider MobileNet-v2 as our base network, but
the results are similar for all the considered models and
configurations (App.~\ref{appx:explaccuracy} and
\ref{appx:additional-uncertainty}). The first step that we can do to
extract insights about the decision process, is to check which
samples in the memory set have positive weights -- the set $M_c \cup M_e$.
\figurename~\ref{fig:mem_example} shows this set ordered by the magnitude of content
weights for four different inputs: each couple shares the same memory set as
additional input, but each set of used samples -- those associated
with a positive weight -- is different. In particular, consider the images in
\figurename~\ref{fig:mem_svhn}, where the only change is a lateral shift made to
center different numbers. Despite their closeness in the input space, samples in
memory are totally different: the first set contains images of ``5'' and ``3'',
while the second set contains mainly images of ``1'' and few images of ``7''. We
can infer that probably the network is focusing on the shape of the number in
the center to classify the image, ignoring colors and the surrounding context.
Conversely, in \figurename~\ref{fig:mem_cifar} the top samples in memory are images
with similar colors and different shapes, telling us that the network is wrongly
focusing on the association between color in the background and color of the
object in the center. This means that the inspection of samples in the
set $M_c \cup M_e$ can give us some insights about the decision process.

Once we have defined the nature of the samples in the memory set that influence
the inference process, we can check whether the content weight
ranking is meaningful for Memory Wrap predictions. To verify that this
is the case, consider the most represented sample inside the memory
vector (i.e. the sample $x^i_{m_k}$ associated with the highest
content weight). Then, let $g(x^i_{m_k})$ be the prediction obtained by
replacing the current input with this sample, and the current memory
set $S_i$ with a new one. If the sample influences in a
significant way the decision process and if it can be considered as a
good proxy for the current prediction $g(x_i)$ (i.e a good explanation
by example), then $g(x^i_{m_k})$ should be equal to $g(x_i)$.
Therefore, we set the \emph{explanation accuracy} as a measure that
checks how many times the sample in the memory set with the highest weight
is predicted in the same class of the current
image. Table~\ref{table:expaccuracy} shows the explanation accuracy of
MobileNet-v2 in all the considered configurations. We observe that
Memory Wrap reaches high accuracy, meaning that the content weights
ranking is reliable. Additionally, its accuracy is very close to
the baseline that uses only the memory, despite the fact this latter
is favored by its design, meaning that the memory content heavily
influences the decision process.

\begin{figure}[t!]
  \centering
  \subcaptionbox{\label{fig:4a}}{\centering \includegraphics[scale=0.32]{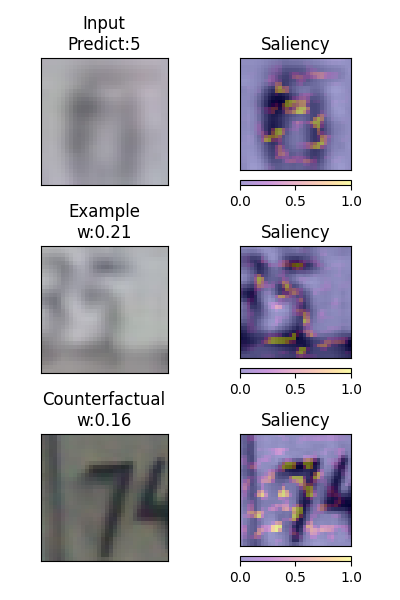}}
  \subcaptionbox{\label{fig:4b}}{\centering \includegraphics[scale=0.32]{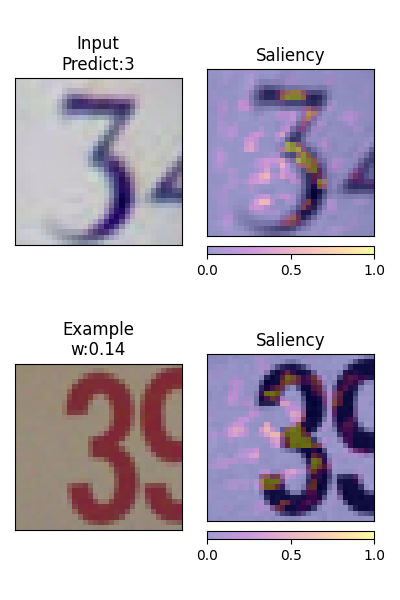}}
  \subcaptionbox{\label{fig:4c}}{\centering \includegraphics[scale=0.32]{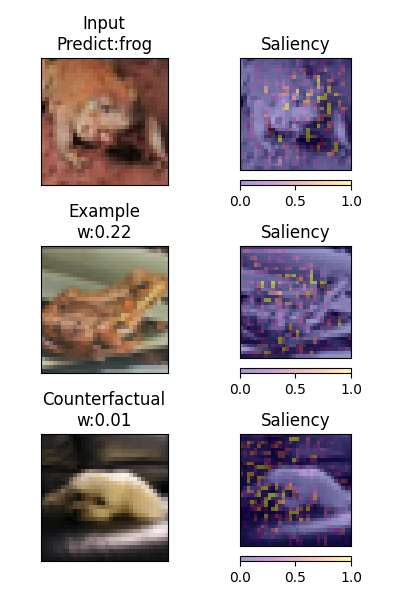}}
  \subcaptionbox{\label{fig:4d}}{\centering \includegraphics[scale=0.32]{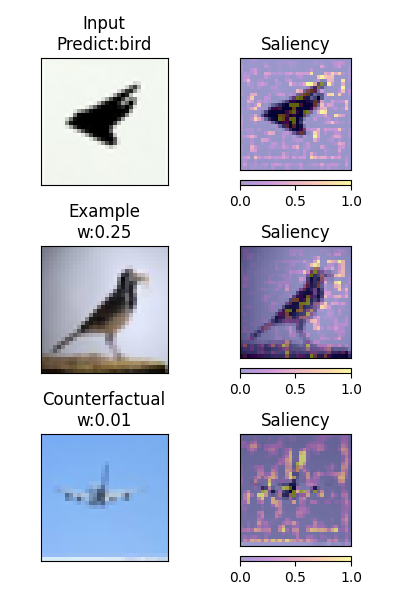}}
  \caption{Integrated Gradients heatmaps of the input, the explanation by example associated
  with the highest weight in memory, and (eventually)
  the counterfactual associated with the highest weight.
  Each heatmap highlights the pixels that have a positive impact towards the
  current prediction.}\label{fig:4}
\end{figure}

Clearly, the same test cannot be applied to counterfactuals, because, by
construction, they are samples of a different class. However, we can inspect
what happens when a counterfactual is the sample with the highest weight. We
find (Table~\ref{table:counteraccuracy}) that the model accuracy is much lower
in these cases, meaning that its predictions are often wrong and one can use
this information to alert the user that the decision process could be
unreliable.

Since the memory is actively used during the inference phase, we can
use attribution methods to extract further insights about the decision
process (see App.~\ref{appx:attributions} for a discussion about the
choice of the attribution method). For example, Figure~\ref{fig:4}
shows heatmaps obtained applying Integrated Gradients\footnote{We use
  the Captum library~\cite{Kokhlikyan2020}}~\cite{Sundararajan2017}, a
method that highlights the most relevant pixels for the current
prediction, exploiting the gradients.
For both \figurename~\ref{fig:4a} and \figurename~\ref{fig:4d}, the model predicts the wrong class. 
In the \ref{fig:4d} case, the heatmap of the explanation by example
tells us that the model focuses on bird and sky colors, ignoring the
unusual shape of the airplane, very different from previously known
shapes for airplanes, which are represented by the counterfactual with
a very low weight and a heatmap that focuses only on the
sky. Conversely in the \ref{fig:4c} case, the model ignores colors and
focuses on the head shape, a feature that is highlighted both in the
input image and in the explanations. Finally, sometimes (see
\figurename~\ref{fig:4b}) counterfactuals are missing, and this means
that the model is sure about its prediction and it uses only examples
of the same class.


%% file: sections/conclusion.tex
\section{Conclusion and future research}
\label{sec:discussion}
In this paper, we presented an extension for neural networks that
allows a more efficient use of the training dataset in
settings where few data are available. Moreover, we propose an
approach to extract explanations based
on similar examples and counterfactuals.\\
Future work could explore the reduction of current limitations, like the
memory space needed to store the memory samples and their gradients
(App.~\ref{appx:cost}). Another limitation is
that the memory mechanism based on similarity could amplify the bias learned by the encoder. As shown in Sect.~\ref{sec:method-explanation}, the identification of such an event is straightforward, but currently there are not countermeasures against it. A new adaptive or algorithmic selection
mechanism of memory samples or a regularization method could mitage the bias and it could improve
the fairness of Memory Wrap. 
Finally, the findings
of this paper open up also possible extensions on different problems like
semi-supervised learning, where the self uncertainty detection of Memory Wrap
could be useful, and domain adaption.

%% file: appendix.tex
\appendix

\section{Appendix}
\label{sec:appx}

\input{sections/setup}
\input{sections/layer}
\input{sections/attributions}
\input{sections/explacc.tex}

\newpage
\input{sections/architectures}

\newpage

\input{sections/cost}

\newpage
\input{sections/impact}
\input{sections/uncertainty}
\input{sections/images}
\clearpage

%% file: sections/setup.tex
\subsection{Training Details}
\label{appx:trainingdetails}
\subsubsection{Datasets.}
We test our approach on image classification tasks using the Street
View House Number (SVHN) dataset~\cite{Netzer2011} (GNU 3.0 license), CINIC10~\cite{Darlow2018}(MIT license) and
CIFAR10~\cite{Krizhevsky2009}(MIT license). SVHN is a dataset containing $\sim$73k
images of house numbers in natural scenarios. The goal is to recognize
the right digit in the image. Sometimes some distracting digits are
present next to the centered digits of interest. CIFAR10 is an
extensively studied dataset containing $\sim$60k images where each
image represents one of the 10 classes of the
dataset. Finally, CINIC10 is relatively new dataset containing $\sim$90k images that tries to bridge
the gap betwen CIFAR10 and ImageNet in terms of difficulty, using the same
classes of CIFAR10 and a subset of merged images from both CIFAR10 and ImageNet.

At the beginning of our experiments, we randomly extract from training
sets a validation test of 6k images for each dataset. The
images are normalized and, in CIFAR10 and CINIC10, we also apply an augmentation
based on random horizontal flips. We do not use the random crop
augmentation because, in some preliminary tests, it can hurt the
performance, as a random crop can often isolate a portion of the image
containing only the background. The memory in this case will
retrieve similar examples based only on the background, pushing the
network to learn useless shortcuts, degrading the
performance.

The subsets of the training dataset to train models with 1000, 2000 and 5000 samples are extracted randomly and change in every run. This means that we extract 15 different subsets of the dataset and then test all the configurations on these subsets. We fixed the seed using the range (0,15) to make the results reproducible.

\subsubsection{Training details.}
The implementation of the architectures for our encoders $f(x)$
starts from the PyTorch
implementations of Kuang
Liu\footnote{https://github.com/kuangliu/pytorch-cifar}. To train the
models, we follow the setup of Huang et al.~\cite{Huang2017}, where
they are trained for 40 epochs in SVHN and 300 epochs in CIFAR10. In
both cases, we apply the Stochastic Gradient Descent (SGD) algorithm
providing a learning rate that starts from 1e-1 and decreases by a
factor of 10 after 50\% and 75\% of epochs.  Note that this
configuration is not optimal neither for baselines nor for Memory Wrap
and you can reach higher performance on both cases by choosing another
set of hyperparameters tuned in each setting. However, this makes
quite fair the comparison across different models and datasets.
We ran our experiment using a cloud hosted NVIDIA A100 and a GTX 3090. 
\paragraph{Memory Set}
Regarding memory samples, in an ideal setting one should provide a new memory
set for each input during the training process, however this makes both
the training and the inference process slowe due to computational limits. We
simplified the process by providing a single memory set for each new batch. The
consequence is that performance at testing/validation time can be influenced by
the batch size used: a high batch size means a high dependency on the random
selection. To limit the instability, we fix a batch size at testing time of 500
and we repeat the test phase 5 times, extracting the average accuracy across
all repetitions.

%% file: sections/layer.tex
\subsection{Final Layer.}
\label{appx:finallayer}
In this section, we describe and motivate the choice of the parameters of the
last layer.  In principle, we can use any function as the last
layer. In some preliminary tests, we compared a linear layer against a
multi-layer perceptron. We found that linear layers require lower
learning rates (in the range of [1e-2,1e-4]) to work well in our
settings. However, for the considered
datasets and models, the standard configuration requires a decreasing learning rate that starts
from high values. To make the comparison fair, we choose, instead, a
multi-layer perceptron that seems more stable and reliable at high
learning rates. The choice of a linear layer is appealing, because it
makes easier the inspection of the contribution of each sample in the
memory to compute the final prediction, and in principle, one could
obtain similar or higher results if hyperparameters are suitably tuned. We use a multi-layer perceptron containing only 1 hidden
layer. The input dimension of such a layer will be clearly
$dim(l_f) = 2dim(e_{x_i})$ being $dim(e_{x_i}) = dim(\mathbf{v_{S_i}})$ for the
Memory Wrap and
$dim(l_f) = dim(\mathbf{v_{S_i}})$, for the baseline that uses only the memory
vector. The size of the hidden layer $dim(h_{l_f})$ is a
hyper-parameter that we fix multiplying the input size by a factor of
2.

%% file: sections/attributions.tex
\subsection{Attribution Methods.}
\label{appx:attributions}
As described in the paper, it is possible to use an attribution method to
highlight the most important pixels for both the input image and the memory
set, with respect to the current prediction. The only requirement is that the
attribution method must support multi-input settings. We use the implementation
of Integrated Gradients~\cite{Sundararajan2017} provided by the Captum
library~\cite{Kokhlikyan2020}. Note that, one of the main problems of these
attribution methods is the choice of the baseline~\cite{Sundararajan2017}: it should
represent the absence of information. In the image domain, it is difficult to
choose the right baseline, because there is a high variability of shapes and
colors. We selected a white image as the baseline, because it is a common
background on SVHN dataset, but this choice generates two effects: 1) it makes
the heatmaps blind to white color and this means, for example, that heatmaps for
white numbers on a black background focus on edges of numbers instead of the
inner parts; 2) it is possible to obtain a different heatmap by changing the
baseline. 

%% file: sections/explacc.tex
\subsection{Explanation Accuracy.}
\label{appx:explaccuracy}

Table ~\ref{table:additional-expaccuracy} shows the complete set of experiments for the computation of the explanation accuracy. 
\begin{table}[ht]
    \centering
    \caption{Mean Explanation accuracy and standard deviation over 15 runs of the
    sample in the memory set with the highest sparse content attention weight.}
    \label{table:additional-expaccuracy}
    \begin{tabular}[t]{@{}rrrr@{}}
      \toprule
      &\multicolumn{2}{c}{\textbf{Explanation Accuracy Memory Wrap \%}}\\
      \midrule
      Dataset&Samples&EfficientNetB0&ResNet18\\
      \midrule
      \quad SVHN&1000 & 86.09 $\pm$ 0.76 & 69.85 $\pm$ 3.29\\
      \quad &2000 &  90.04  $\pm$ 0.51 & 84.64 $\pm$ 2.02\\
      \quad &5000 &  93.19 $\pm$ 0.49 & 92.45 $\pm$ 0.26\\
      \midrule
      \quad CIFAR10&1000 & 80.06 $\pm$ 0.85 & 72.95 $\pm$ 1.00\\
      \quad &2000 & 80.27 $\pm$ 0.57 & 78.42 $\pm$ 0.58\\
      \quad &5000& 85.05 $\pm$ 0.47 & 85.86 $\pm$ 0.46\\
      \bottomrule

    \end{tabular}
\end{table}

%% file: sections/architectures.tex
\subsection{Additional Architectures.}
\label{appx:architectures}
Table~\ref{table:additional-SVHN} and Table~\ref{table:additional-CIFAR10} show
the performance of GoogLeNet~\cite{Szegedy2015}, DenseNet~\cite{Huang2017}, and
ShuffleNet~\cite{Zhang2018} on both datasets. We can observe that the performance trend follows that of the other architectures.
\begin{table*}[h]
    \centering
    \caption{Avg. accuracy and standard deviation over 15 runs of the baselines and Memory Wrap, when the training dataset is a subset of SVHN. For each configuration, we highlight in bold the best result and results that are within its margin.}
    \label{table:additional-SVHN}
    \begin{tabular}{@{}rrrrrr@{}}
      \toprule
      \multicolumn{5}{c}{\textbf{Reduced SVHN Avg. Accuracy\%}}\\
      \midrule
       &&\multicolumn{3}{c}{\textbf{Samples}}\\
      Model&Variant&1000&2000&5000\\
      \midrule
      \quad GoogLeNet  & Standard  & 25.25 $\pm$ 9.39  & 61.45 $\pm$ 16.56  &  88.63 $\pm$ 2.60 \\
      \quad &Only Memory     & \textbf{66.35 $\pm$ 6.93}  & 87.10 $\pm$ 1.17  & 92.16 $\pm$ 0.28 \\
      \quad &Memory Wrap    & \textbf{74.66 $\pm$ 9.01}  & \textbf{88.32 $\pm$ 0.78}  & \textbf{92.52 $\pm$ 0.25} \\
      \midrule
      \quad DenseNet  &  Standard   & 60.93 $\pm$ 9.21  & 83.47 $\pm$ 1.16  &  89.39 $\pm$ 0.60 \\
      \quad & Only Memory      &  40.94 $\pm$ 12.06  & 79.12 $\pm$ 5.36  &  \textbf{89.69 $\pm$ 0.63} \\
      \quad & Memory Wrap      &  \textbf{73.69 $\pm$ 4.20}  & \textbf{85.12 $\pm$ 0.62}  & \textbf{90.07 $\pm$ 0.49} \\
      \midrule
      \quad ShuffleNet  & Standard  & 27.09 $\pm$  6.05 & 60.05 $\pm$ 8.76  &  83.19 $\pm$ 1.00 \\
      \quad & Only Memory      & \textbf{32.22 $\pm$ 3.47}  & 60.06 $\pm$ 3.32 &  \textbf{85.56 $\pm$ 0.60} \\
      \quad & Memory Wrap     & \textbf{33.60 $\pm$ 4.69}  & \textbf{67.35 $\pm$ 3.19}  & \textbf{85.04 $\pm$ 0.74} \\
      \bottomrule
    \end{tabular}
  
  \end{table*}

  \begin{table*}[ht]
      \centering
      \caption{Avg. accuracy and standard deviation over 15 runs of the baselines and Memory Wrap, when the training dataset is a subset of CIFAR10. For each configuration, we highlight in bold the best result and results that are within its margin.}
      \label{table:additional-CIFAR10}
      \begin{tabular}{@{}rrrrrr@{}}
        \toprule
        \multicolumn{5}{c}{\textbf{Reduced CIFAR10 Avg. Accuracy\%}}\\
        \midrule
         &&\multicolumn{3}{c}{\textbf{Samples}}\\
        Model&Variant&1000&2000&5000\\
        \midrule
        \quad GoogLeNet  & Standard  & 51.91 $\pm$ 3.14  & 63.90 $\pm$ 2.21  &  79.09 $\pm$ 1.28 \\
        \quad &Only Memory    & 54.25 $\pm$ 0.80  & 66.00 $\pm$ 1.27  &  79.65 $\pm$ 0.59 \\
        \quad &Memory Wrap    & \textbf{55.91 $\pm$ 1.20}  & \textbf{66.79 $\pm$ 1.03}  &  \textbf{80.27 $\pm$ 0.49} \\
        \midrule
        \quad DenseNet  &  Standard   & \textbf{46.99 $\pm$  1.61} & \textbf{56.95 $\pm$ 1.68}  &  73.72 $\pm$ 1.41 \\
        \quad & Only Memory      & \textbf{46.20 $\pm$ 1.47}  & \textbf{58.16 $\pm$   1.82} &  \textbf{75.77 $\pm$ 1.31} \\
        \quad & Memory Wrap      & \textbf{47.64 $\pm$  1.58} & \textbf{58.60 $\pm$  1.85} &  \textbf{75.50 $\pm$ 1.33} \\
        \midrule
        \quad ShuffleNet  & Standard  & \textbf{37.86 $\pm$ 1.16}  & 45.85 $\pm$ 1.26  &  65.92 $\pm$ 1.54 \\
        \quad & Only Memory      & \textbf{38.15 $\pm$ 1.14}  & \textbf{48.91 $\pm$ 2.12}  &  70.05 $\pm$ 0.84 \\
        \quad & Memory Wrap     & \textbf{37.90 $\pm$ 1.15}  & \textbf{47.50 $\pm$ 1.79}  &  \textbf{68.52 $\pm$ 1.38} \\
        \bottomrule
      \end{tabular}
    
    \end{table*}

    \begin{table*}[ht]
        \centering
        \caption{Avg. accuracy and standard deviation over 15 runs of the baselines and Memory Wrap, when the training dataset is a subset of CINIC10. For each configuration, we highlight in bold the best result and results that are within its margin.}
        \label{table:additional-CINIC10}
        \begin{tabular}{@{}rrrrrr@{}}
          \toprule
          \multicolumn{5}{c}{\textbf{Reduced CINIC10 Avg. Accuracy\%}}\\
          \midrule
           &&\multicolumn{3}{c}{\textbf{Samples}}\\
          Model&Variant&1000&2000&5000\\
          \midrule
          \quad GoogLeNet  & Standard  & 38.97 $\pm$ 1.16  & 47.83 $\pm$ 1.09  &  \textbf{58.47 $\pm$ 0.91}\\
          \quad &Only Memory    & 40.77 $\pm$ 0.78  & 48.53 $\pm$ 1.05  &     57.86 $\pm$ 0.55\\
          \quad &Memory Wrap    & \textbf{42.19 $\pm$ 0.92}  & \textbf{50.47 $\pm$ 0.77}  &  \textbf{58.98 $\pm$ 0.68}\\
          \midrule
          \quad DenseNet  &  Standard   & \textbf{36.33 $\pm$ 0.84}  & 41.78 $\pm$ 0.92  & 52.63  $\pm$ 0.95\\
          \quad & Only Memory      & 35.64 $\pm$  1.18 & 42.77 $\pm$ 0.69  &   \textbf{54.16 $\pm$ 0.58}\\
          \quad & Memory Wrap    & \textbf{37.02 $\pm$ 0.95}  & \textbf{43.55 $\pm$ 1.05}  & 53.59 $\pm$ 0.61\\
          \midrule
          \quad ShuffleNet  & Standard  & \textbf{28.32 $\pm$ 0.85}  & 33.49 $\pm$ 0.93  &  46.36 $\pm$ 1.03\\
          \quad & Only Memory      & \textbf{28.68 $\pm$ 0.93}  & \textbf{35.33 $\pm$ 1.09} & \textbf{48.25  $\pm$ 0.90}\\
          \quad & Memory Wrap     & \textbf{28.94 $\pm$ 1.06}  & \textbf{34.30 $\pm$ 0.85}  &  47.33 $\pm$ 1.34\\
          \bottomrule
        \end{tabular}
      
      \end{table*}

%% file: sections/cost.tex
\subsection{Computational Cost}
\label{appx:cost}
In this section, we describe briefly the changes in the computational cost when adding the Memory Wrap.
\subsubsection{Parameters.}
 The network size’s increment depends mainly on the output dimensions of the encoder and on the choice of the final layer. In Table \ref{table:parameters} we examine the case of an MLP as the final layer and MobileNet, ResNet18, or EfficientNet as the encoder. We replace a linear layer of dim $(a,b)$ with a MLP with 2 layers of dimension $(a,a \times 2)$ and $(a \times 2,b)$ passing from a*b parameters to $a \times (a \times 2) + (a \times 2) \times b$. So the increment is mainly caused by the $a$ parameter. A possible solution to reduce the number of parameters would be to add a linear layer between the encoder and the Memory Wrap that projects data in a lower dimensional space, preserving the performance as much as possible.

\begin{table}[ht]
  \setlength{\tabcolsep}{4pt}
    \centering
    \caption{Number of parameters for the models with and without Memory Wrap. The column \textit{dimension} indicates the number of output units of the encoder.}
    \label{table:parameters}
    \begin{tabular}[t]{@{}rrr@{}}
      \toprule
      \multicolumn{3}{c}{\textbf{Number of parameters}}\\
      \midrule
      Model& Dimension &Parameters\\
      \midrule
      \quad EfficientNetB0 & 320 & 3 599 686  \\
      \quad  Only Memory & - & 3 808 326  \\
      \quad  Memory Wrap & - & 4 429 766  \\
      \midrule
      \quad MobileNet-v2 & 1280 & 2 296 922  \\
      \quad  Only Memory & - & 5 589 082  \\
      \quad  Memory Wrap  & - & 15 447 642
      \\
      \midrule
      \quad ResNet18 & 512 & 11 173 962  \\
      \quad  Only Memory & - &  11 704 394 \\
      \quad  Memory Wrap  & - & 13 288 522  \\
      \bottomrule
    \end{tabular}
  \end{table}

\subsubsection{Space complexity}
Regarding the space required for the memory, in principle, we should provide a new memory set for each input during the training process. Let be $m$ the size of memory and n the dimension of the batch, the new input will contain $m \times n $ samples in place of $n$. For large batch sizes and a large number of samples in memory, this cost can be too high. To reduce its memory footprint, we simplified the process by providing a single memory set for each new batch, maintaining the space required to a more manageable $m+n$. 

\subsubsection{Time Complexity}
Time complexity depends on the
number of training samples included in the memory set. In our
experiments we used 100 training samples for each step as a trade-off
between performance and training time, doubling the
training time due to the added gradients and the additional encoding
of the memory set. However, in the inference phase, we can obtain
nearly the same time complexity by fixing the memory set a priori and
computing its encodings only the first time.

%% file: sections/impact.tex
\subsection{Impact of Memory Size}
\label{appx:memorysize}
The memory size is one of the hyper-parameters of Memory Wrap. We chose empirically a value (100) that is a trade-off between the number of samples for each class (10), the minimum number of samples considered in the training set (1000), the training time and the performance. The value is motivated by the fact that we want enough samples for each class to get more representative samples for that class, but, at the same time, we don’t want that often the current sample is also included in the memory set and the architecture exploits this fact. \\
Increasing the number of samples can increase the performance too (Table~\ref{table:impactSVHN}), but it comes at the cost of training and inference time.
For example, an epoch of EfficientNetB0, trained using 5000 samples, lasts \textasciitilde 9 seconds when the memory contains 20 samples, \textasciitilde 16 seconds when the memory contains 300 samples and \textasciitilde 22 seconds when the memory contains 500 samples. 

\begin{table*}[h!]
    \centering
    \caption{Avg. accuracy and standard deviation over 5 runs of the configuration of Memory Wrap trained using different number of samples in memory, when the training dataset is a subset of SVHN.}
    \label{table:impactSVHN}
    \begin{tabular}[h]{@{}rrrrrr@{}}
      \toprule
      \multicolumn{5}{c}{\textbf{Reduced SVHN Avg. Accuracy\%}}\\
      \midrule
       &&\multicolumn{3}{c}{\textbf{Samples}}\\
      Model&Samples&1000&2000&5000\\
      \midrule
      \quad EfficientNetB0  & 20 & 64.95 $\pm$ 2.46 & 75.80 $\pm$ 1.17    &   84.86  $\pm$ 0.99   \\
      \quad & 100  & 67.16 $\pm$ 1.33  & 77.02 $\pm$  2.20 & 85.82 $\pm$ 0.45\\
      \quad & 300  & 66.70 $\pm$ 1.58  & 77.97 $\pm$ 1.34  &   85.37  $\pm$ 0.68 \\
      \quad &  500 & 66.76 $\pm$  0.98 & 77.67 $\pm$ 1.17  & 85.25 $\pm$ 1.02\\
      \midrule
      \quad MobileNet & 20  & 63.42 $\pm$  2.46 & 80.92 $\pm$ 1.42  &  88.33 $\pm$ 0.36 \\
      \quad & 100  & 68.31 $\pm$ 1.53  & 81.28 $\pm$ 0.69  &  88.47 $\pm$ 0.10 \\
      \quad &  300 & 65.08 $\pm$ 0.30  & 82.05 $\pm$ 0.75  & 88.93  $\pm$ 0.37 \\
      \quad &  500 & 69.88 $\pm$ 1.76  & 80.92 $\pm$  1.74 & 88.61  $\pm$ 0.32 \\
      \midrule
      \quad ResNet18 &  20 & 39.32 $\pm$ 7.21  & 72.54 $\pm$ 3.03  & 87.30  $\pm$ 0.41 \\
      \quad & 100  & 40.38 $\pm$ 9.32  & 74.36 $\pm$ 2.69  &  87.39 $\pm$ 0.45 \\
      \quad & 300 & 44.42 $\pm$ 10.97  & 74.63 $\pm$ 3.28  & 87.75  $\pm$ 0.62 \\
      \quad &  500 & 40.59 $\pm$ 12.27  & 76.97 $\pm$ 2.48  &  87.55 $\pm$ 0.35 \\
      \bottomrule
    \end{tabular}
  
  \end{table*}

%% file: sections/uncertainty.tex
\subsection{Uncertainty Detection.}
\label{appx:additional-uncertainty}
Table ~\ref{table:additional-uncertainty} shows the accuracy reached by the
models on inputs where the sample in memory associated with the
highest weight is a counterfactual. In these cases, models seem
unsure about their predictions, making a lot of mistakes with
respect to classical settings. This behavior can be observed on
$\sim$10\% of the testing dataset.
  \begin{table}[h!]
      \centering
      \caption{Accuracy reached by the model on images where the sample with the
      highest weight in memory set is a counterfactual. The accuracy is computed as
      the mean over 15 runs.}
      \label{table:additional-uncertainty}
      \begin{tabular}[t]{@{}rrrr@{}}
        \toprule
        &\multicolumn{2}{c}{\textbf{Explanation Accuracy Memory Wrap \%}}\\
        \midrule
        Dataset&Samples&EfficientNetB0&ResNet18\\
        \midrule
        \quad SVHN&1000 & 37.10 $\pm$ 1.45 & 31.10 $\pm$ 4.65\\
        \quad &2000 & 40.77 $\pm$ 2.80 & 46.23 $\pm$ 1.74\\
        \quad &5000 & 46.94 $\pm$ 1.49 & 50.04 $\pm$ 0.96\\
        \midrule
        \quad CIFAR10&1000 & 29.37 $\pm$ 0.84 & 31.74 $\pm$ 0.95\\
        \quad &2000 & 35.44 $\pm$ 0.98 & 35.67 $\pm$ 1.25\\
        \quad &5000 & 45.47 $\pm$ 0.90 & 42.17 $\pm$ 0.93\\
        \bottomrule
  
      \end{tabular}
  \end{table}

   \cleardoublepage

%% file: sections/images.tex
\subsection{Additional Memory Images.}
\label{appx:memimages}
\begin{figure}[!ht]
  \begin{subfigure}{.5\columnwidth}
    \centering
    \includegraphics[width=.5\columnwidth]{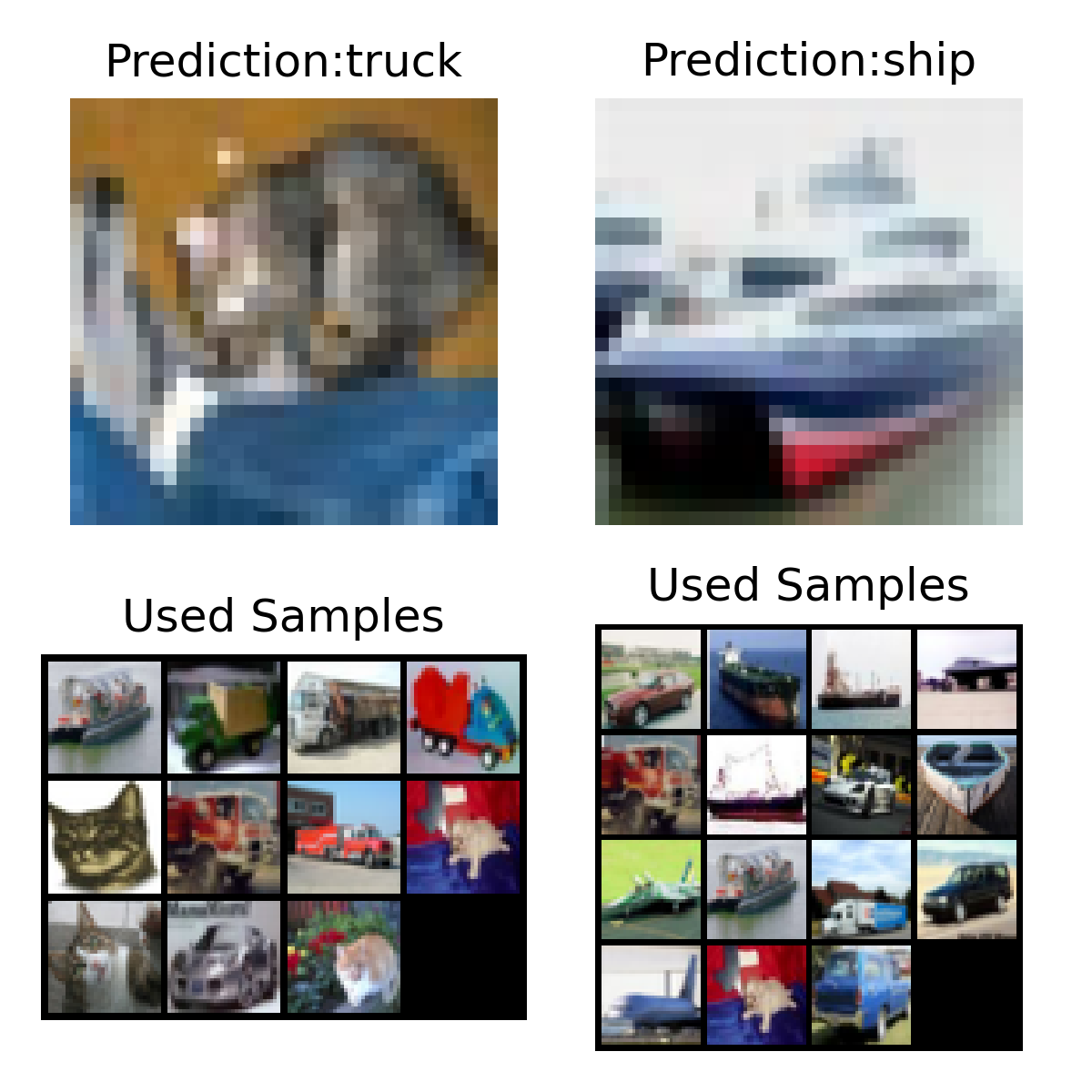}
    \caption{}
    
  \end{subfigure}%
  \begin{subfigure}{.5\columnwidth}
    \centering
    \includegraphics[width=.5\columnwidth]{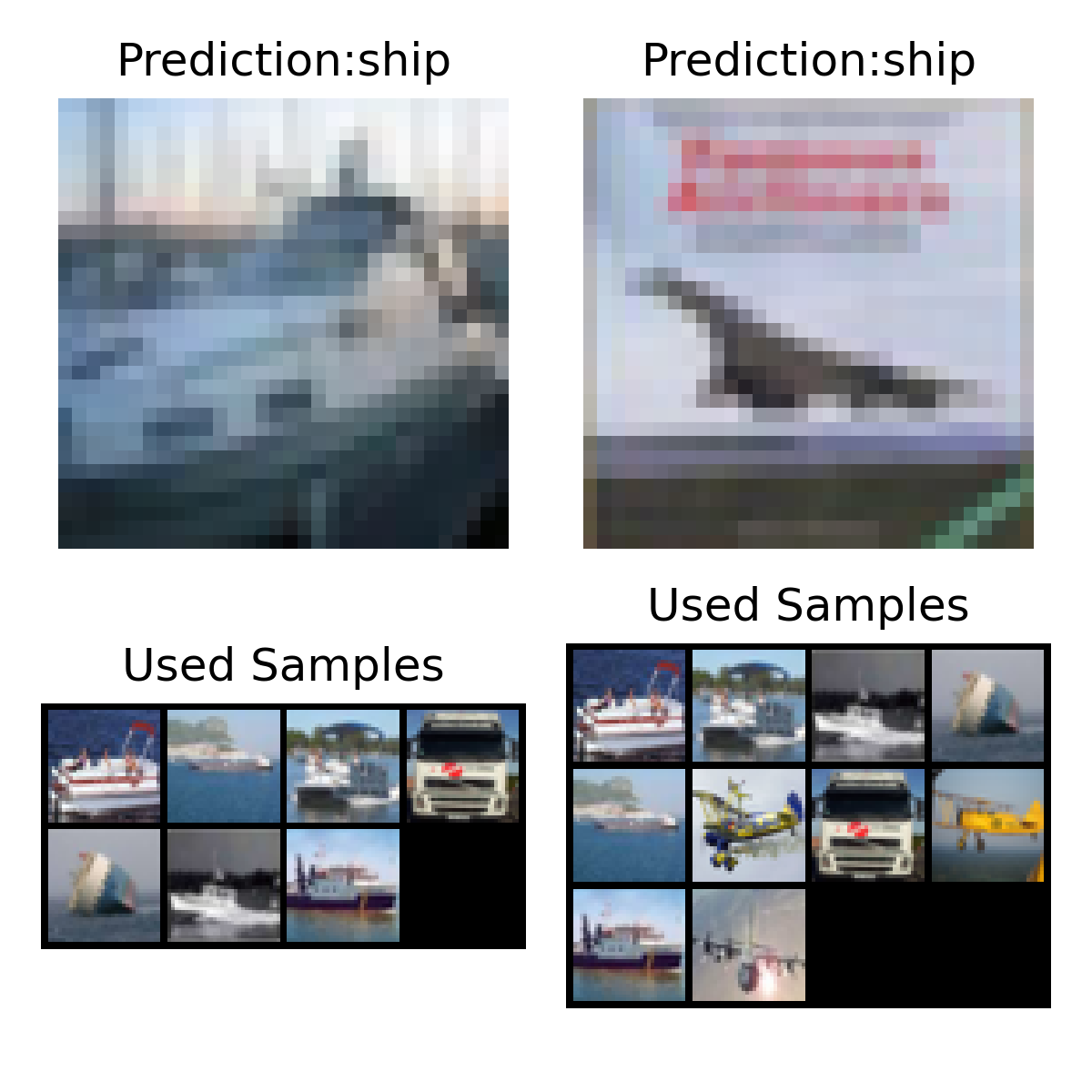}
    \caption{}
    
  \end{subfigure}

  \begin{subfigure}{.5\columnwidth}
    \centering
    \includegraphics[width=.5\columnwidth]{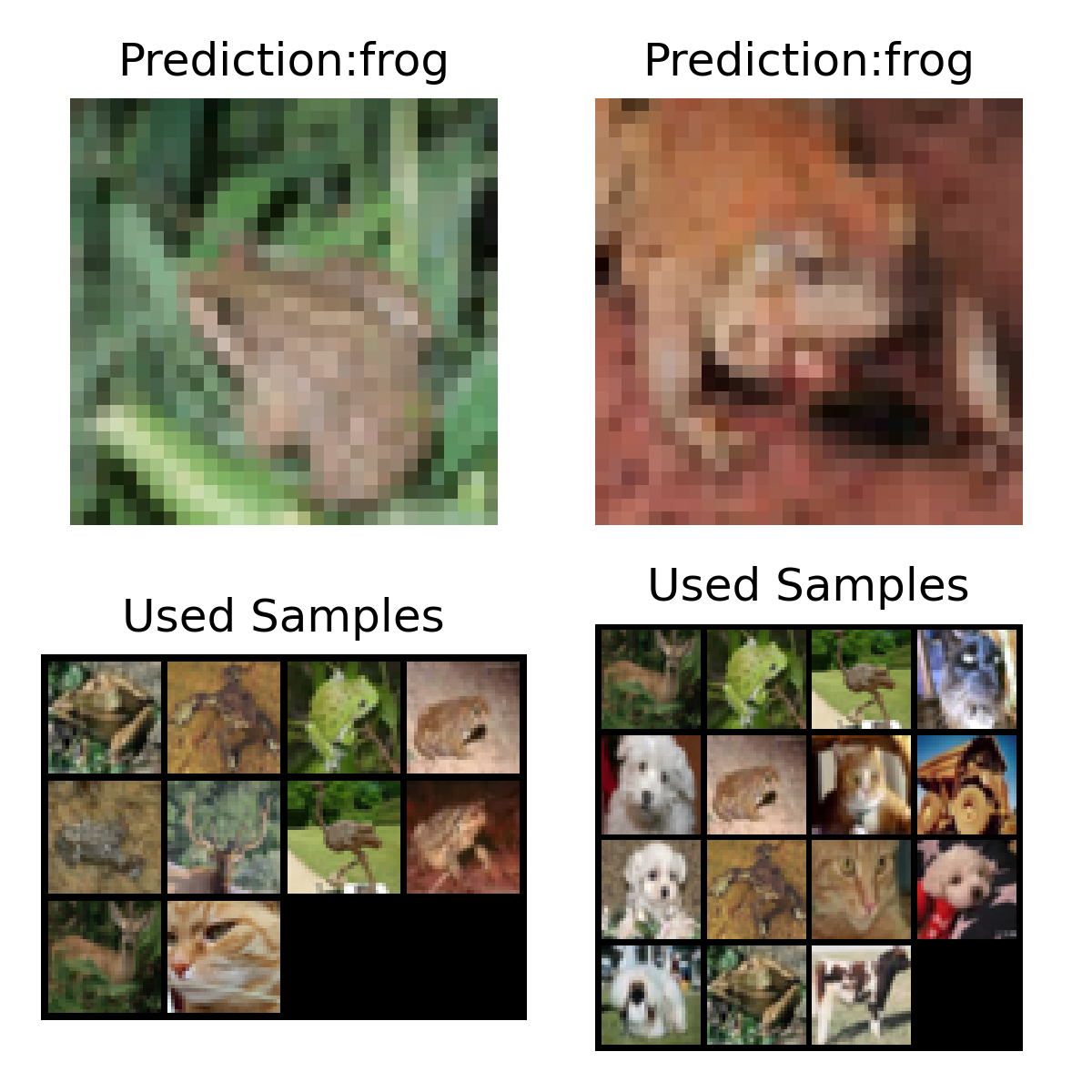}
    \caption{}
    
  \end{subfigure}
  \begin{subfigure}{.5\columnwidth}
    \centering
    \includegraphics[width=.5\columnwidth]{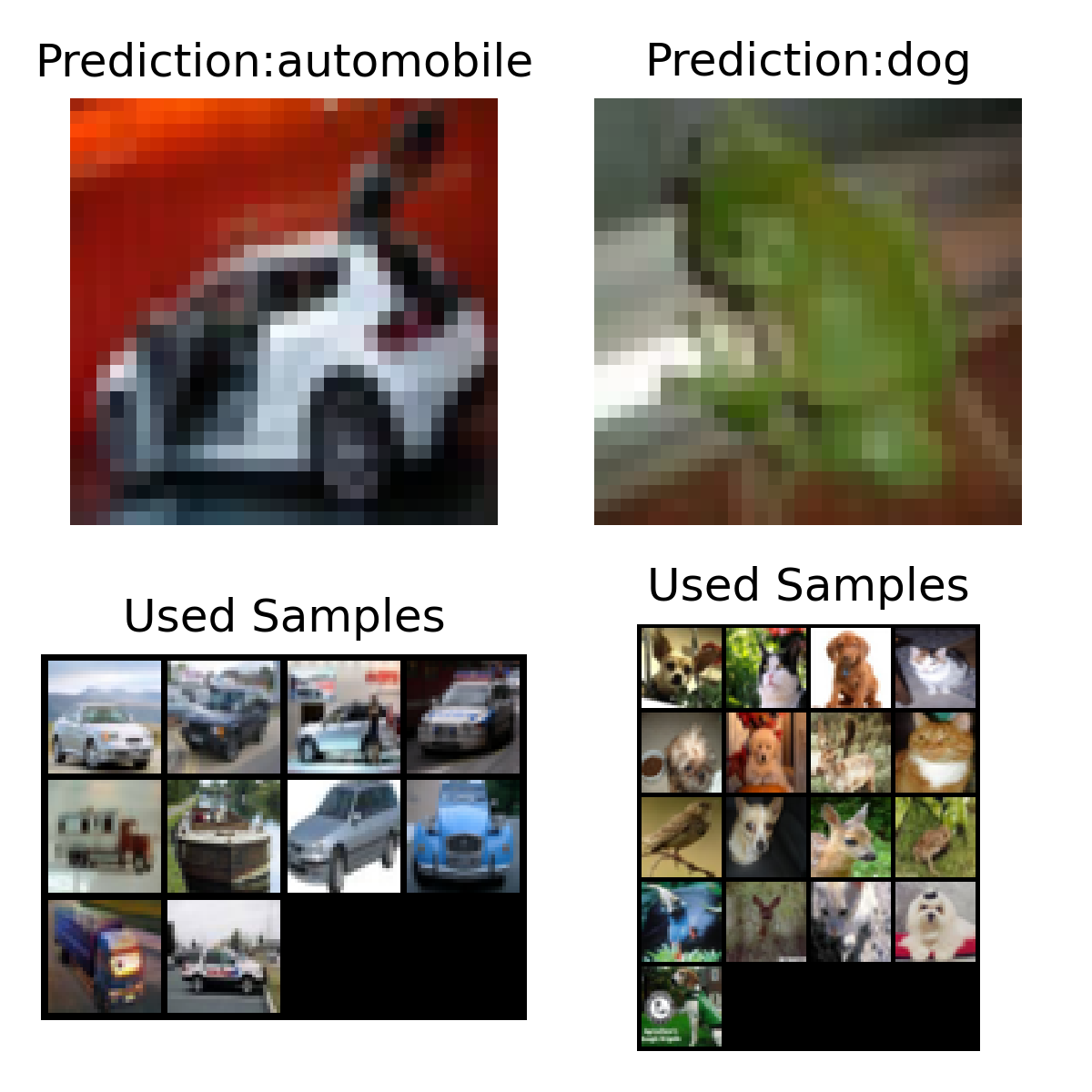}
    \caption{}
    
  \end{subfigure}

  \caption{Inputs from CIFAR10 dataset (first rows), their associated predictions, and an overview of the samples in memory that have an active influence on the decision process -- i.e. the samples from where the memory vector is built -- (second row).}
\end{figure}

\begin{figure}[!ht]
  \begin{subfigure}{.5\columnwidth}
    \centering
    \includegraphics[width=.5\columnwidth]{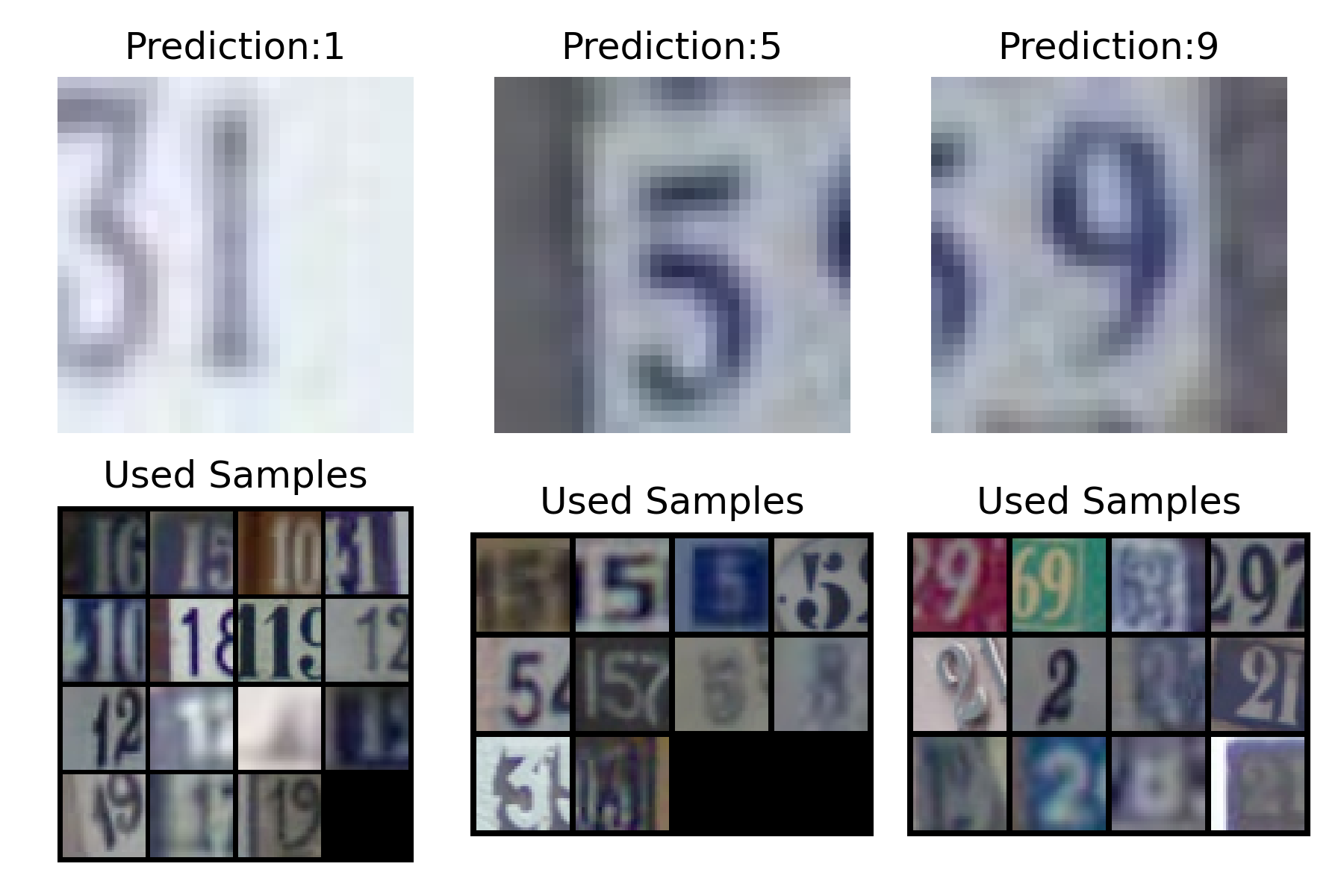}
    \caption{}
    
  \end{subfigure}%
  \begin{subfigure}{.5\columnwidth}
    \centering
    \includegraphics[width=.5\columnwidth]{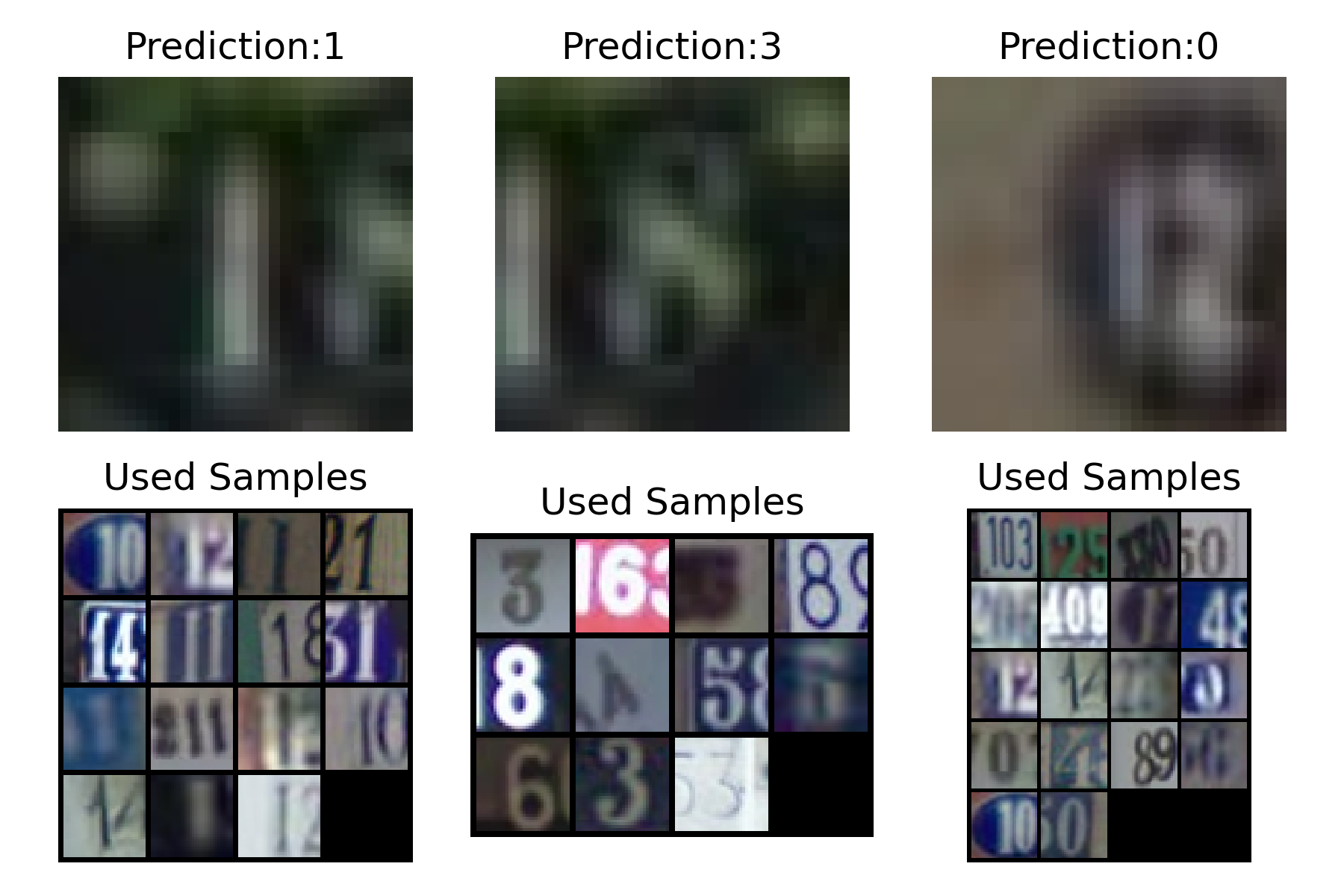}\caption{}
    
  \end{subfigure}

  \begin{subfigure}{.5\columnwidth}
    \centering
    \includegraphics[width=.5\columnwidth]{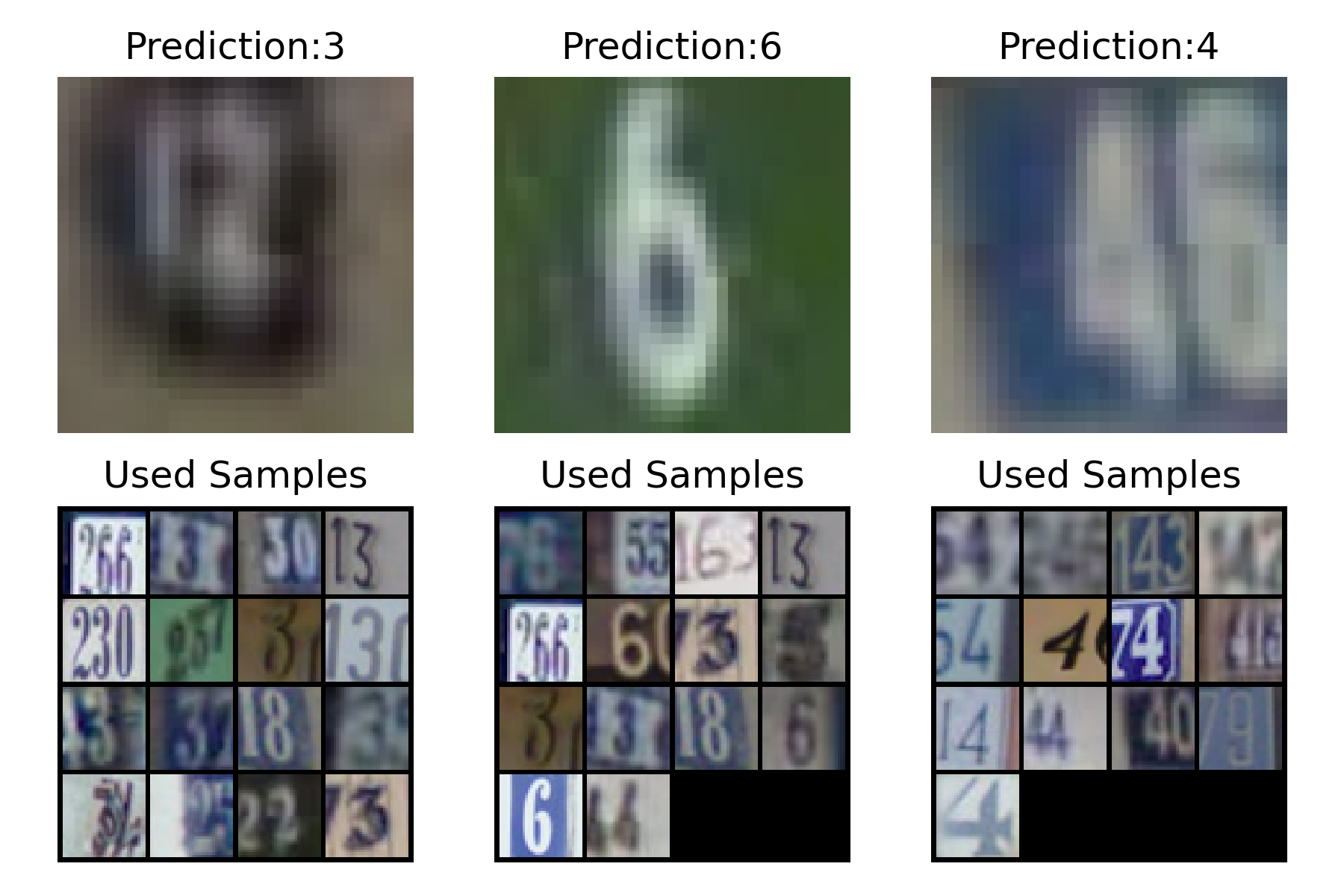}\caption{}
    
  \end{subfigure}
  \begin{subfigure}{.5\columnwidth}
    \centering
    \includegraphics[width=.5\columnwidth]{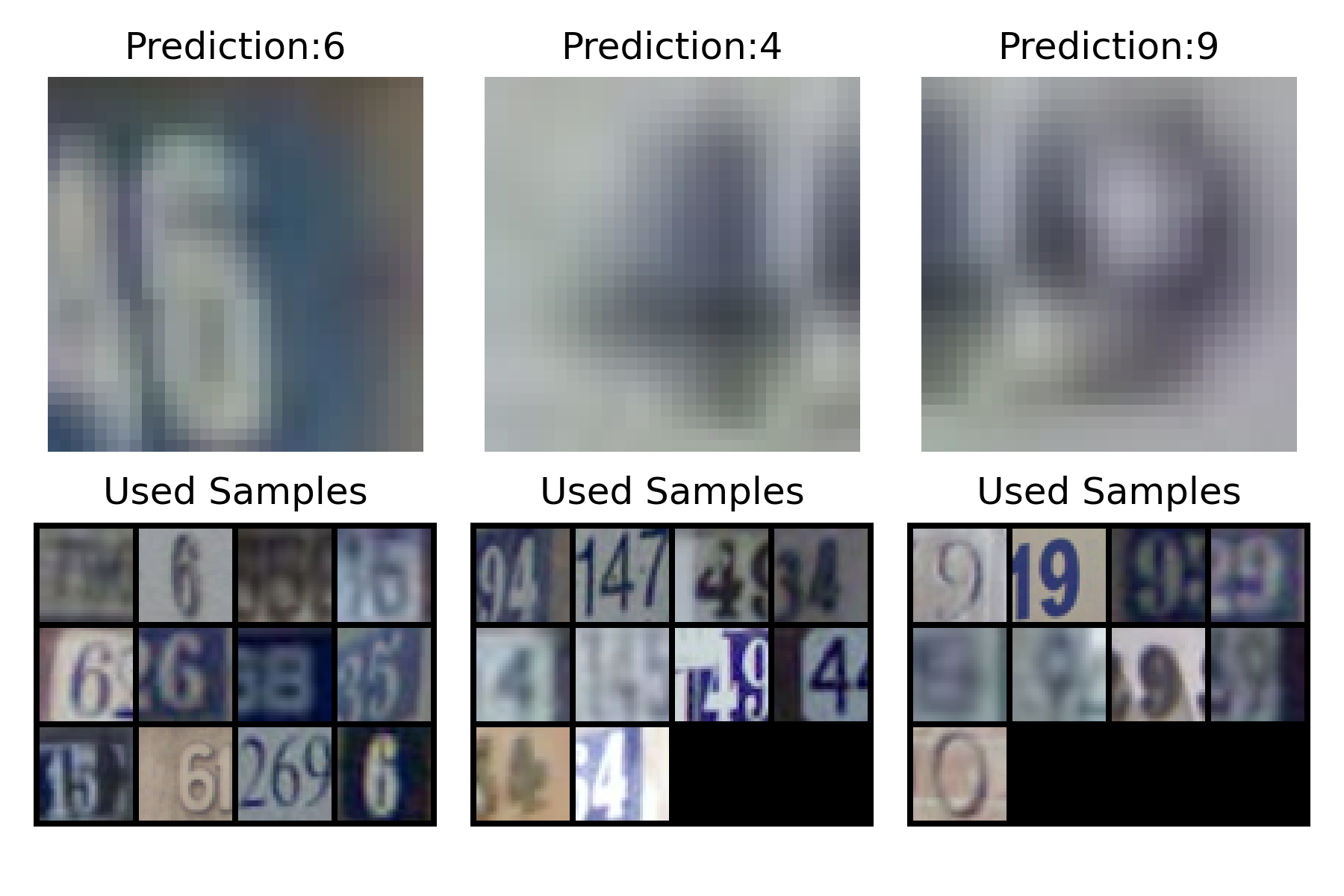}\caption{}
    
  \end{subfigure}

  \caption{Inputs from SVHN dataset (first rows), their associated predictions, and an overview of the samples in memory that have an active influence on the decision process -- i.e. the samples from where the memory vector is built -- (second row).}
\end{figure}

\begin{figure}[!ht]
  \begin{subfigure}{.5\columnwidth}
    \centering
    \includegraphics[width=.5\columnwidth]{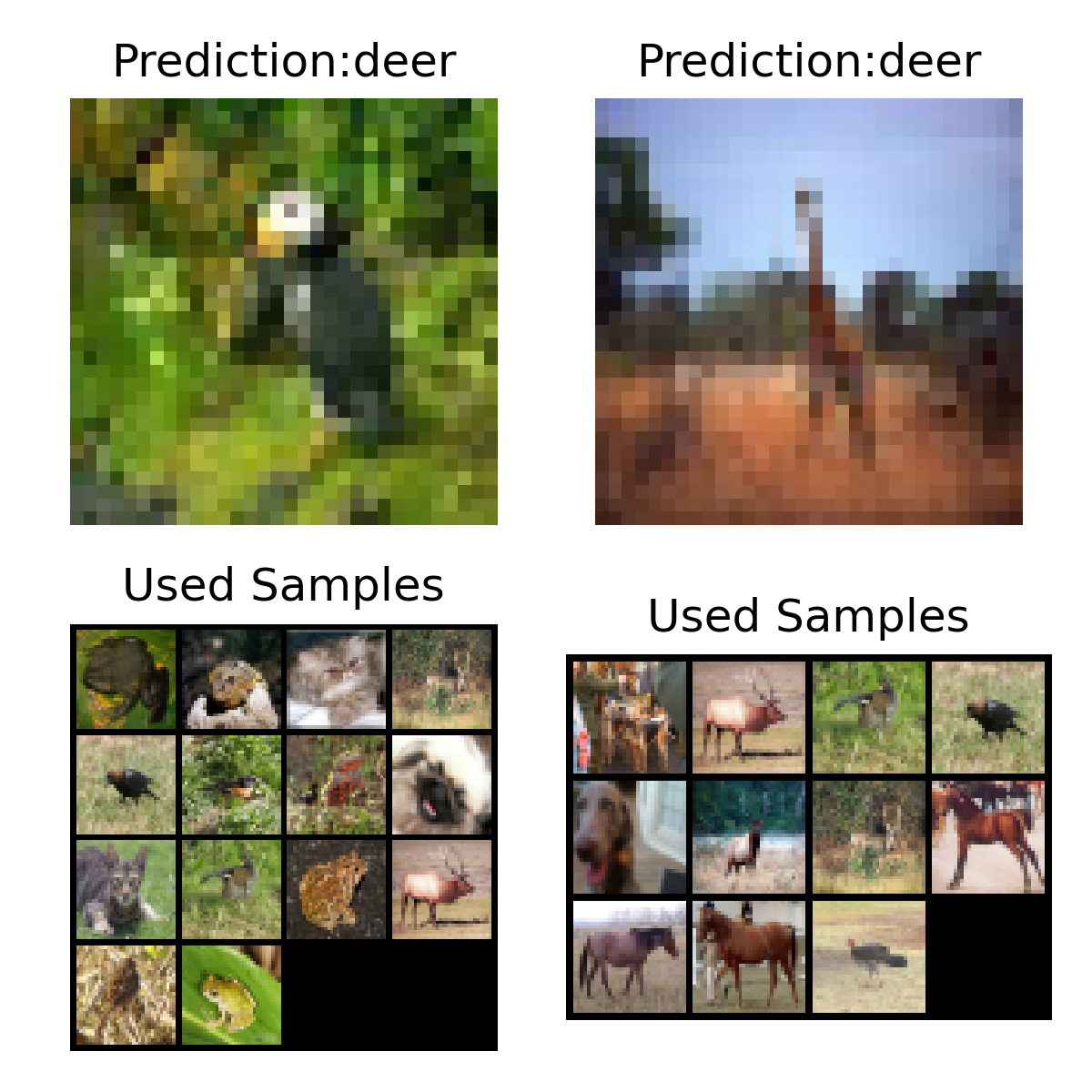}
    \caption{}
    
  \end{subfigure}%
  \begin{subfigure}{.5\columnwidth}
    \centering
    \includegraphics[width=.5\columnwidth]{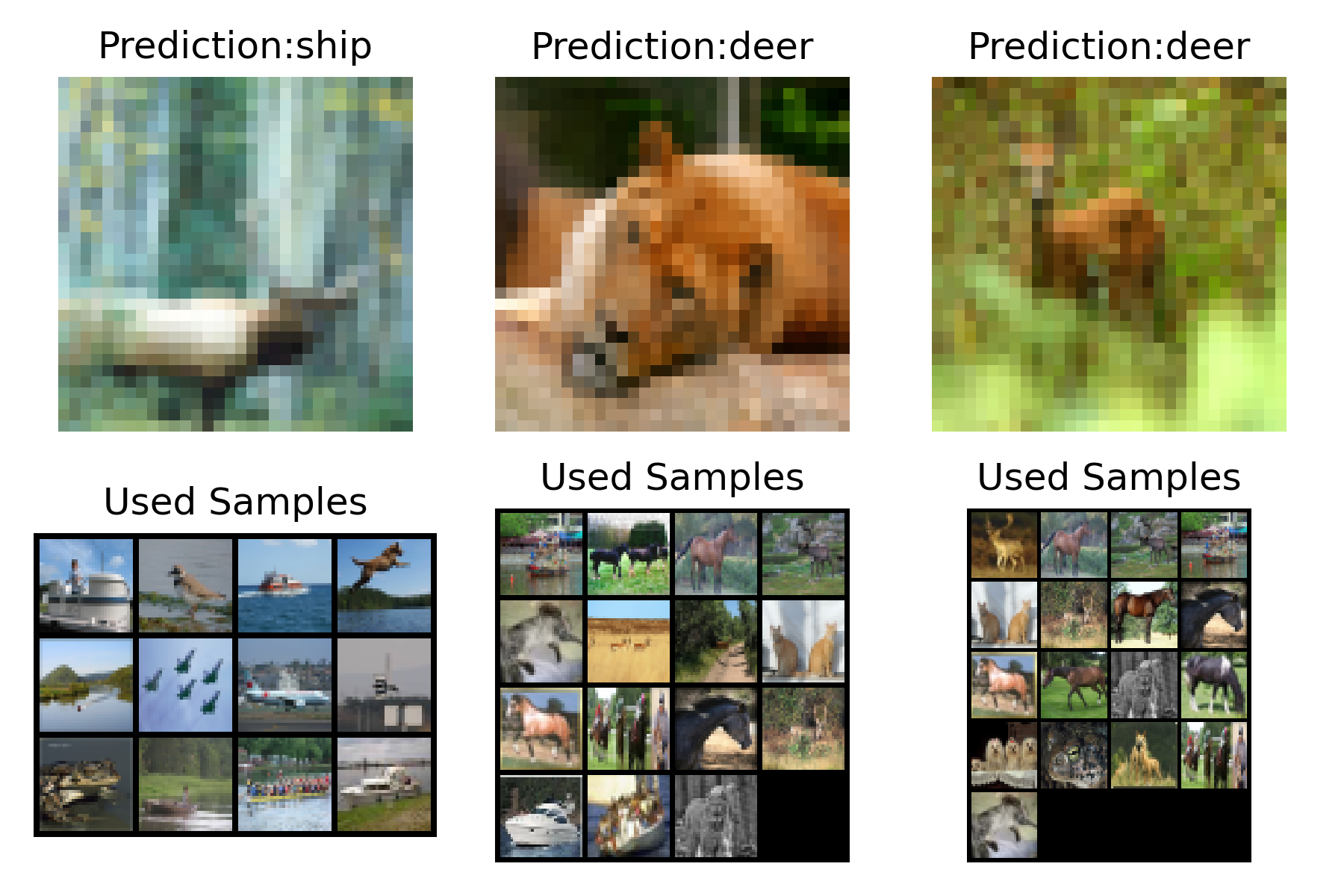}\caption{}
    
  \end{subfigure}

  \begin{subfigure}{.5\columnwidth}
    \centering
    \includegraphics[width=.5\columnwidth]{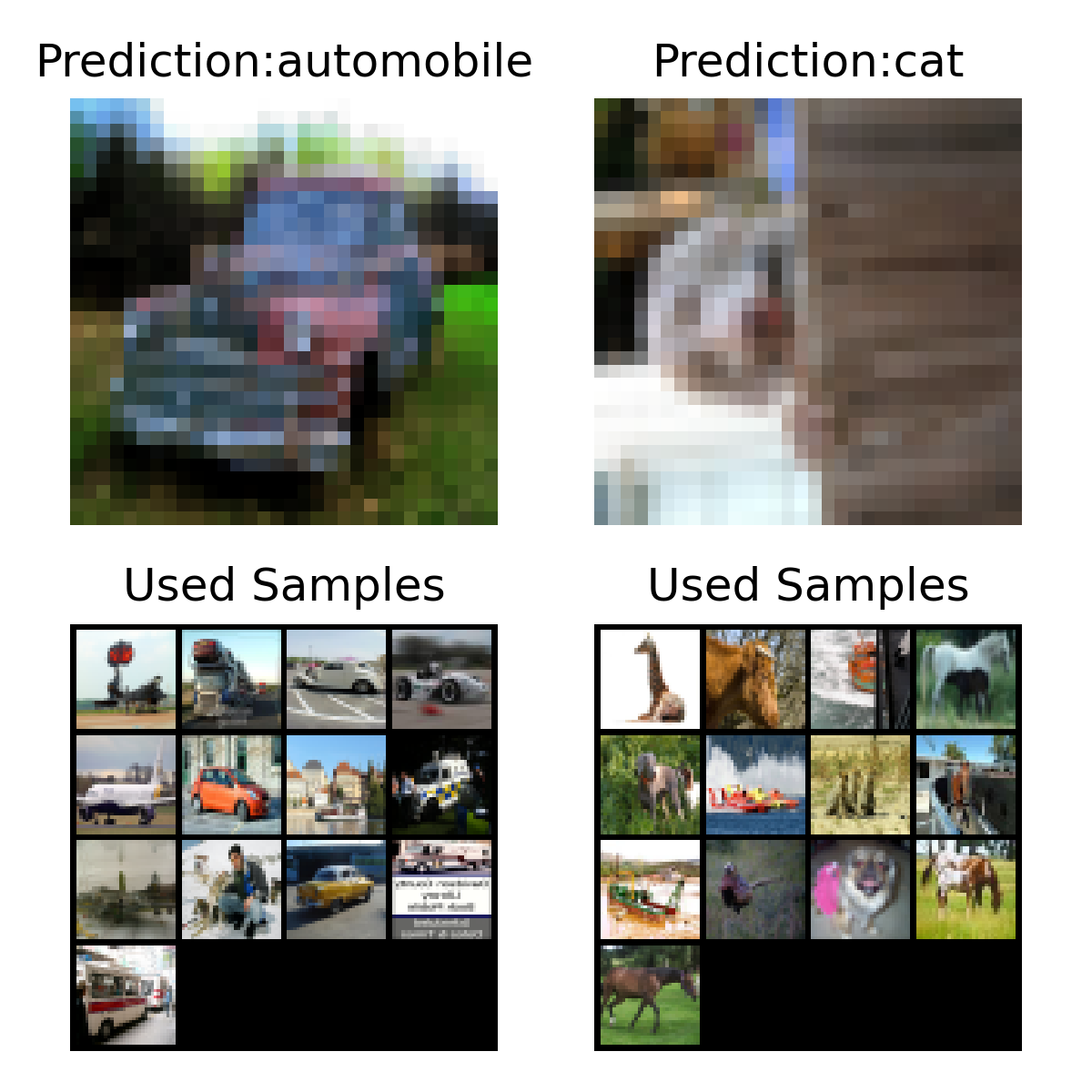}\caption{}
    
  \end{subfigure}
  \begin{subfigure}{.5\columnwidth}
    \centering
    \includegraphics[width=.5\columnwidth]{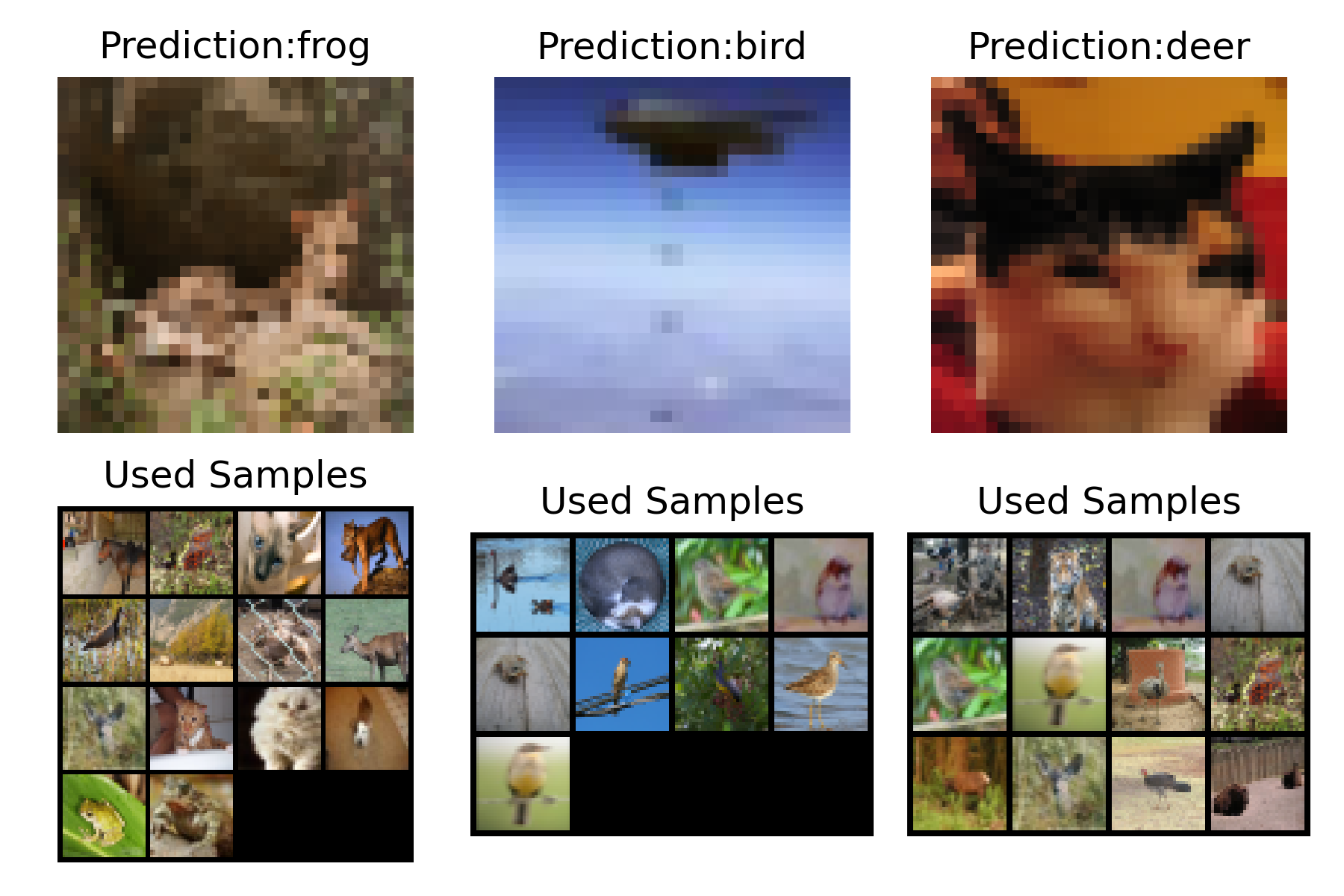}\caption{}
    
  \end{subfigure}

  \caption{Inputs from CINIC10 dataset (first rows), their associated predictions, and an overview of the samples in memory that have an active influence on the decision process -- i.e. the samples from where the memory vector is built -- (second row).}
\end{figure}

\clearpage
\subsection{Additional Heatmaps.}
\begin{figure}[!ht]
  \begin{subfigure}{.5\columnwidth}
    \centering
    \includegraphics[width=.5\columnwidth]{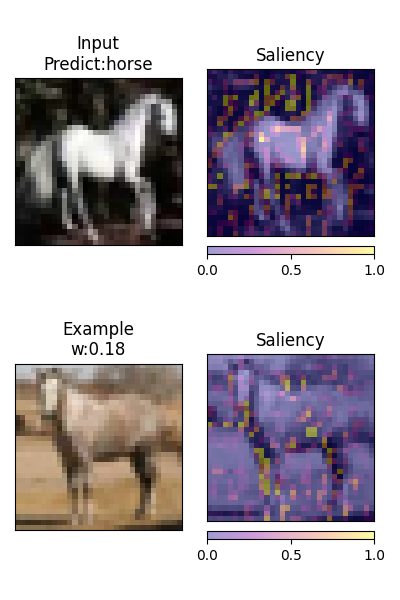}
    \caption{}
    
  \end{subfigure}%
  \begin{subfigure}{.5\columnwidth}
    \centering
    \includegraphics[width=.5\columnwidth]{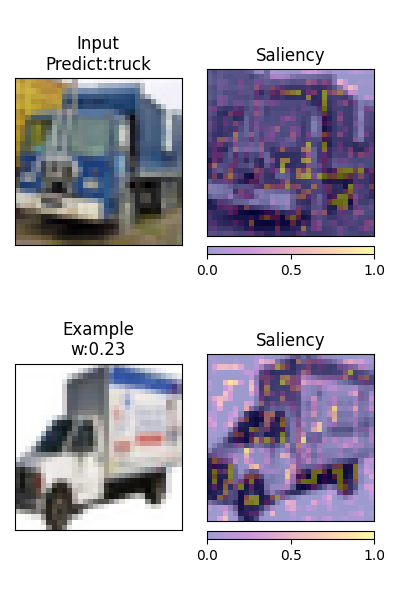}\caption{}
    
  \end{subfigure}

  \begin{subfigure}{.5\columnwidth}
    \centering
    \includegraphics[width=.5\columnwidth]{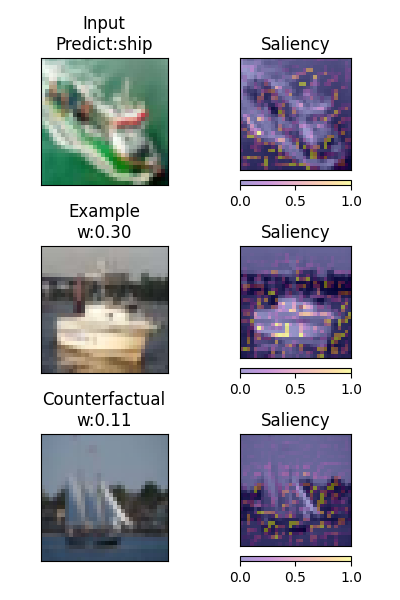}\caption{}
    
  \end{subfigure}
  \begin{subfigure}{.5\columnwidth}
    \centering
    \includegraphics[width=.5\columnwidth]{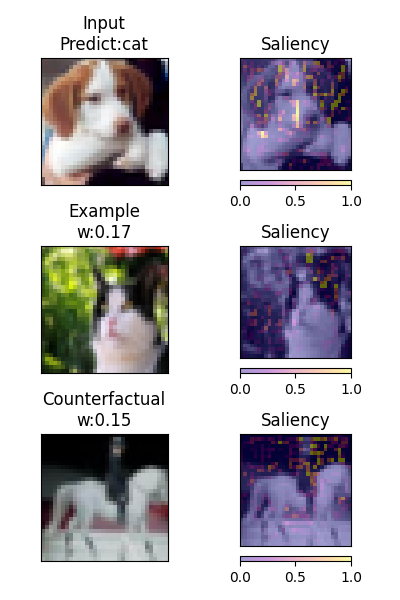}\caption{}
    
  \end{subfigure}
   \caption{Heatmaps computed by the Integrated Gradients method for both the current input and the most relevant samples in memory on the CIFAR10 dataset.}
\end{figure}

\begin{figure}[!ht]
  \begin{subfigure}{.5\columnwidth}
    \centering
    \includegraphics[width=.5\columnwidth]{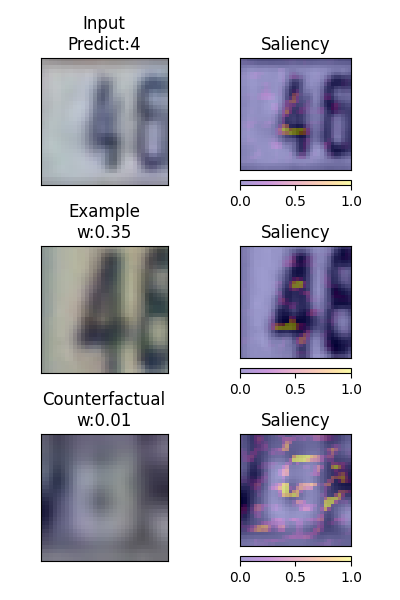}
    \caption{}
    
  \end{subfigure}%
  \begin{subfigure}{.5\columnwidth}
    \centering
    \includegraphics[width=.5\columnwidth]{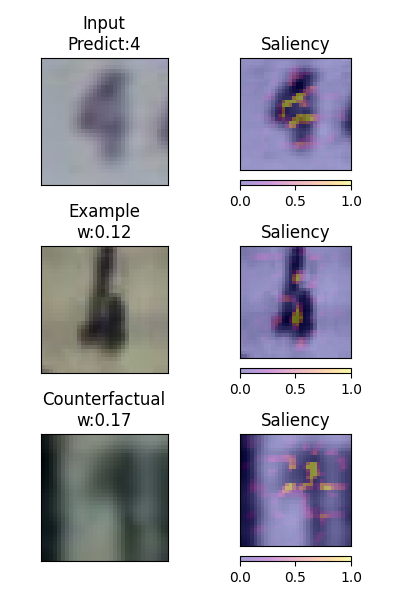}\caption{}
    
  \end{subfigure}

  \begin{subfigure}{.5\columnwidth}
    \centering
    \includegraphics[width=.5\columnwidth]{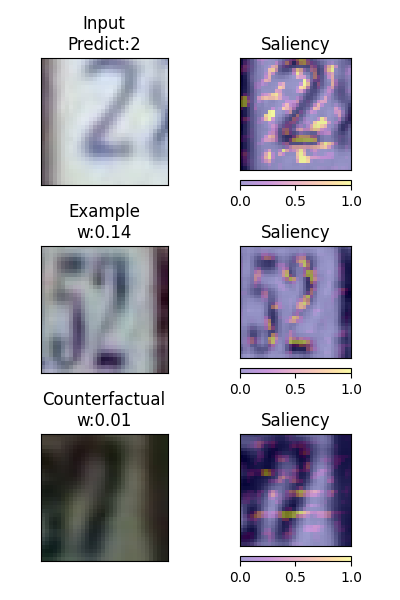}\caption{}
    
  \end{subfigure}
  \begin{subfigure}{.5\columnwidth}
    \centering
    \includegraphics[width=.5\columnwidth]{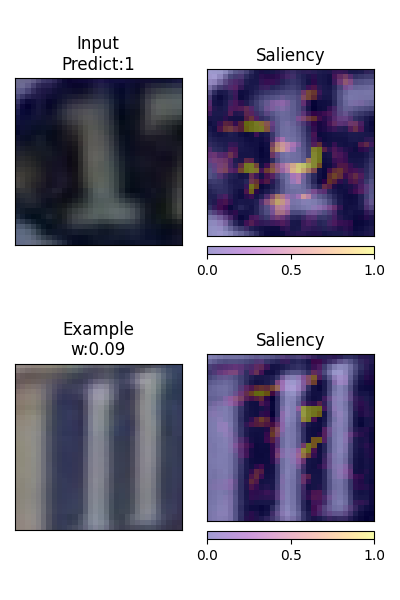}\caption{}
    
  \end{subfigure}
   \caption{Heatmaps computed by the Integrated Gradients method for both the current input and the most relevant samples in memory on the SVHN dataset.}
\end{figure}

\begin{figure}[!ht]
  \begin{subfigure}{.5\columnwidth}
    \centering
    \includegraphics[width=.5\columnwidth]{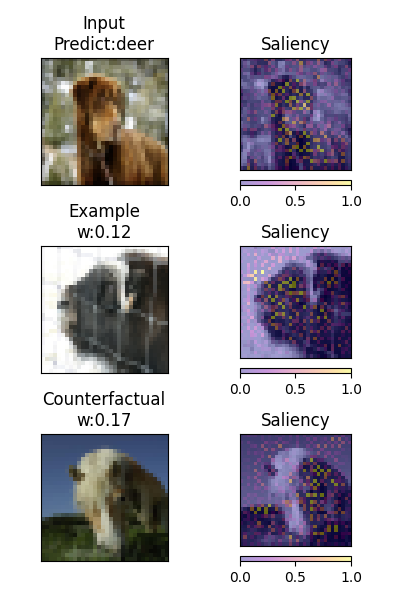}
    \caption{}
    
  \end{subfigure}%
  \begin{subfigure}{.5\columnwidth}
    \centering
    \includegraphics[width=.5\columnwidth]{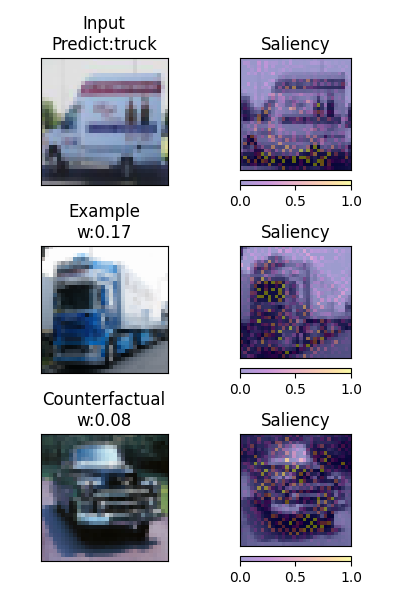}\caption{}
    
  \end{subfigure}

  \begin{subfigure}{.5\columnwidth}
    \centering
    \includegraphics[width=.5\columnwidth]{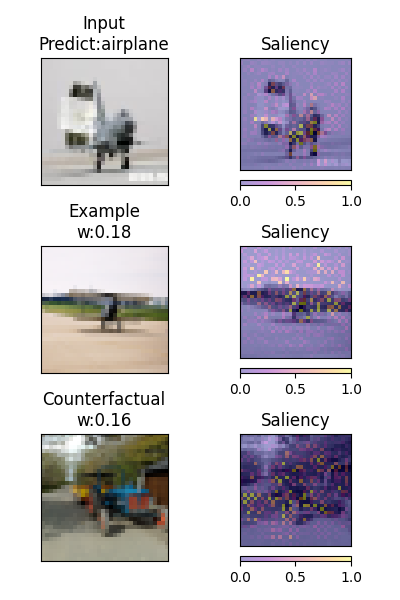}\caption{}
    
  \end{subfigure}
  \begin{subfigure}{.5\columnwidth}
    \centering
    \includegraphics[width=.5\columnwidth]{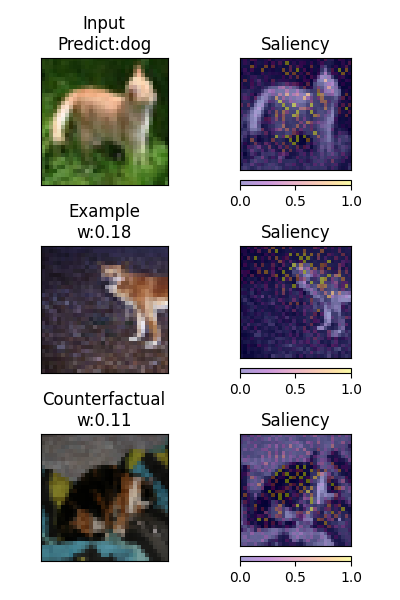}\caption{}
    
  \end{subfigure}
   \caption{Heatmaps computed by the Integrated Gradients method for both the current input and the most relevant samples in memory on the CINIC10 dataset.}
\end{figure}

%% file: arxiv.bbl
\begin{thebibliography}{}

\bibitem[\protect\citeauthoryear{Belle and Papantonis}{2020}]{Belle2020}
Vaishak Belle and Ioannis Papantonis.
\newblock Principles and practice of explainable machine learning.
\newblock 2020.

\bibitem[\protect\citeauthoryear{Cai \bgroup \em et al.\egroup
  }{2018}]{Cai2018}
Qi~Cai, Yingwei Pan, Ting Yao, Chenggang Yan, and Tao Mei.
\newblock Memory matching networks for one-shot image recognition.
\newblock In {\em 2018 {IEEE}/{CVF} Conference on Computer Vision and Pattern
  Recognition}. {IEEE}, 2018.

\bibitem[\protect\citeauthoryear{Cortes and Vapnik}{1995}]{Cortes1995}
Corinna Cortes and Vladimir Vapnik.
\newblock Support-vector networks.
\newblock {\em Machine Learning}, 20(3):273--297, 1995.

\bibitem[\protect\citeauthoryear{Darlow \bgroup \em et al.\egroup
  }{2018}]{Darlow2018}
Luke~N. Darlow, Elliot~J. Crowley, Antreas Antoniou, and Amos~J. Storkey.
\newblock Cinic-10 is not imagenet or cifar-10.
\newblock October 2018.

\bibitem[\protect\citeauthoryear{Dinan \bgroup \em et al.\egroup
  }{2018}]{Dinan2018}
Emily Dinan, Stephen Roller, Kurt Shuster, Angela Fan, Michael Auli, and Jason
  Weston.
\newblock Wizard of wikipedia: Knowledge-powered conversational agents.
\newblock 2018.

\bibitem[\protect\citeauthoryear{Graves \bgroup \em et al.\egroup
  }{2014}]{Graves2014}
Alex Graves, Greg Wayne, and Ivo Danihelka.
\newblock Neural turing machines.
\newblock 2014.

\bibitem[\protect\citeauthoryear{Graves \bgroup \em et al.\egroup
  }{2016}]{Graves2016}
Alex Graves, Greg Wayne, Malcolm Reynolds, Tim Harley, Ivo Danihelka, Agnieszka
  Grabska-Barwi{\'{n}}ska, Sergio~G{\'{o}}mez Colmenarejo, Edward Grefenstette,
  Tiago Ramalho, John Agapiou, Adri{\`{a}}~Puigdom{\`{e}}nech Badia,
  Karl~Moritz Hermann, Yori Zwols, Georg Ostrovski, Adam Cain, Helen King,
  Christopher Summerfield, Phil Blunsom, Koray Kavukcuoglu, and Demis Hassabis.
\newblock Hybrid computing using a neural network with dynamic external memory.
\newblock {\em Nature}, 538(7626):471--476, 2016.

\bibitem[\protect\citeauthoryear{He \bgroup \em et al.\egroup }{2016}]{He2016}
Kaiming He, Xiangyu Zhang, Shaoqing Ren, and Jian Sun.
\newblock Deep residual learning for image recognition.
\newblock In {\em 2016 {IEEE} Conference on Computer Vision and Pattern
  Recognition ({CVPR})}. {IEEE}, 2016.

\bibitem[\protect\citeauthoryear{Howard \bgroup \em et al.\egroup
  }{2017}]{Howard2017}
Andrew~G. Howard, Menglong Zhu, Bo~Chen, Dmitry Kalenichenko, Weijun Wang,
  Tobias Weyand, Marco Andreetto, and Hartwig Adam.
\newblock Mobilenets: Efficient convolutional neural networks for mobile vision
  applications.
\newblock 2017.

\bibitem[\protect\citeauthoryear{Huang \bgroup \em et al.\egroup
  }{2017}]{Huang2017}
Gao Huang, Zhuang Liu, Laurens Van~Der Maaten, and Kilian~Q. Weinberger.
\newblock Densely connected convolutional networks.
\newblock In {\em 2017 {IEEE} Conference on Computer Vision and Pattern
  Recognition ({CVPR})}. {IEEE}, 2017.

\bibitem[\protect\citeauthoryear{Kenny and Keane}{2019}]{Kenny2019}
Eoin~M. Kenny and Mark~T. Keane.
\newblock Twin-systems to explain artificial neural networks using case-based
  reasoning: Comparative tests of feature-weighting methods in {ANN}-{CBR}
  twins for {XAI}.
\newblock In {\em Proceedings of the Twenty-Eighth International Joint
  Conference on Artificial Intelligence}. International Joint Conferences on
  Artificial Intelligence Organization, 2019.

\bibitem[\protect\citeauthoryear{Kokhlikyan \bgroup \em et al.\egroup
  }{2020}]{Kokhlikyan2020}
Narine Kokhlikyan, Vivek Miglani, Miguel Martin, Edward Wang, Bilal Alsallakh,
  Jonathan Reynolds, Alexander Melnikov, Natalia Kliushkina, Carlos Araya, Siqi
  Yan, and Orion Reblitz-Richardson.
\newblock Captum: A unified and generic model interpretability library for
  pytorch.
\newblock 2020.

\bibitem[\protect\citeauthoryear{Krizhevsky}{2009}]{Krizhevsky2009}
A.~Krizhevsky.
\newblock Learning multiple layers of features from tiny images.
\newblock 2009.

\bibitem[\protect\citeauthoryear{Kumar \bgroup \em et al.\egroup
  }{2016}]{Kumar2016}
Ankit Kumar, Ozan Irsoy, Peter Ondruska, Mohit Iyyer, James Bradbury, Ishaan
  Gulrajani, Victor Zhong, Romain Paulus, and Richard Socher.
\newblock Ask me anything: Dynamic memory networks for natural language
  processing.
\newblock In {\em Proceedings of the 33rd International Conference on
  International Conference on Machine Learning}, volume~48 of {\em ICML'16},
  page 1378–1387. JMLR.org, 2016.

\bibitem[\protect\citeauthoryear{La~Rosa \bgroup \em et al.\egroup
  }{2020}]{LaRosa2020}
Biagio La~Rosa, Roberto Capobianco, and Daniele Nardi.
\newblock Explainable inference on sequential data via memory-tracking.
\newblock In {\em Proceedings of the Twenty-Ninth International Joint
  Conference on Artificial Intelligence}. International Joint Conferences on
  Artificial Intelligence Organization, 2020.

\bibitem[\protect\citeauthoryear{Lipton}{2018}]{Lipton2018}
Zachary~C. Lipton.
\newblock The mythos of model interpretability: In machine learning, the
  concept of interpretability is both important and slippery.
\newblock {\em Queue}, 16(3):31–57, 2018.

\bibitem[\protect\citeauthoryear{Liu \bgroup \em et al.\egroup
  }{2019}]{Liu2019}
Shusen Liu, Bhavya Kailkhura, Donald Loveland, and Yong Han.
\newblock Generative counterfactual introspection for explainable deep
  learning.
\newblock 2019.

\bibitem[\protect\citeauthoryear{Looveren and Klaise}{2019}]{Looveren2019}
Arnaud~Van Looveren and Janis Klaise.
\newblock Interpretable counterfactual explanations guided by prototypes.
\newblock 2019.

\bibitem[\protect\citeauthoryear{Ma \bgroup \em et al.\egroup }{2018}]{Ma2018}
Chao Ma, Chunhua Shen, Anthony Dick, Qi~Wu, Peng Wang, Anton van~den Hengel,
  and Ian Reid.
\newblock Visual question answering with memory-augmented networks.
\newblock In {\em 2018 {IEEE}/{CVF} Conference on Computer Vision and Pattern
  Recognition}. {IEEE}, 2018.

\bibitem[\protect\citeauthoryear{Mahajan \bgroup \em et al.\egroup
  }{2019}]{Mahajan2019}
Divyat Mahajan, Chenhao Tan, and Amit Sharma.
\newblock Preserving causal constraints in counterfactual explanations for
  machine learning classifiers.
\newblock 2019.

\bibitem[\protect\citeauthoryear{Netzer \bgroup \em et al.\egroup
  }{2011}]{Netzer2011}
Yuval Netzer, Tao Wang, Adam Coates, Alessandro Bissacco, Bo~Wu, and Andrew~Y.
  Ng.
\newblock Reading digits in natural images with unsupervised feature learning.
\newblock In {\em NIPS Workshop on Deep Learning and Unsupervised Feature
  Learning 2011}, 2011.

\bibitem[\protect\citeauthoryear{Peters \bgroup \em et al.\egroup
  }{2019}]{Peters2019}
Ben Peters, Vlad Niculae, and Andr{\'{e}} F.~T. Martins.
\newblock Sparse sequence-to-sequence models.
\newblock In {\em Proceedings of the 57th Annual Meeting of the Association for
  Computational Linguistics}. Association for Computational Linguistics, 2019.

\bibitem[\protect\citeauthoryear{Santoro \bgroup \em et al.\egroup
  }{2016}]{Santoro2016}
Adam Santoro, Sergey Bartunov, Matthew Botvinick, Daan Wierstra, and Timothy
  Lillicrap.
\newblock Meta-learning with memory-augmented neural networks.
\newblock volume~48 of {\em Proceedings of Machine Learning Research}, pages
  1842--1850. PMLR, 2016.

\bibitem[\protect\citeauthoryear{Snell \bgroup \em et al.\egroup
  }{2017}]{Snell2017}
Jake Snell, Kevin Swersky, and Richard Zemel.
\newblock Prototypical networks for few-shot learning.
\newblock In {\em Proceedings of the 31st International Conference on Neural
  Information Processing Systems}, NIPS'17, page 4080–4090. Curran Associates
  Inc., 2017.

\bibitem[\protect\citeauthoryear{Sukhbaatar \bgroup \em et al.\egroup
  }{2015}]{Sukhbaatar2015}
Sainbayar Sukhbaatar, Arthur Szlam, Jason Weston, and Rob Fergus.
\newblock End-to-end memory networks.
\newblock In {\em Proceedings of the 28th International Conference on Neural
  Information Processing Systems - Volume 2}, NIPS'15, page 2440–2448. MIT
  Press, 2015.

\bibitem[\protect\citeauthoryear{Sundararajan \bgroup \em et al.\egroup
  }{2017}]{Sundararajan2017}
Mukund Sundararajan, Ankur Taly, and Qiqi Yan.
\newblock Axiomatic attribution for deep networks.
\newblock In {\em Proceedings of the 34th International Conference on Machine
  Learning - Volume 70}, ICML'17, page 3319–3328. JMLR.org, 2017.

\bibitem[\protect\citeauthoryear{Szegedy \bgroup \em et al.\egroup
  }{2015}]{Szegedy2015}
Christian Szegedy, Wei Liu, Yangqing Jia, Pierre Sermanet, Scott Reed, Dragomir
  Anguelov, Dumitru Erhan, Vincent Vanhoucke, and Andrew Rabinovich.
\newblock Going deeper with convolutions.
\newblock In {\em Proceedings of the IEEE Conference on Computer Vision and
  Pattern Recognition (CVPR)}, June 2015.

\bibitem[\protect\citeauthoryear{Tan and Le}{2019}]{Tan2019}
Mingxing Tan and Quoc Le.
\newblock {E}fficient{N}et: Rethinking model scaling for convolutional neural
  networks.
\newblock volume~97 of {\em Proceedings of Machine Learning Research}, pages
  6105--6114. PMLR, 2019.

\bibitem[\protect\citeauthoryear{Vinyals \bgroup \em et al.\egroup
  }{2016}]{Vinyals2016}
Oriol Vinyals, Charles Blundell, Timothy Lillicrap, Koray Kavukcuoglu, and Daan
  Wierstra.
\newblock Matching networks for one shot learning.
\newblock In {\em Proceedings of the 30th International Conference on Neural
  Information Processing Systems}, NIPS'16, page 3637–3645. Curran Associates
  Inc., 2016.

\bibitem[\protect\citeauthoryear{Wachter \bgroup \em et al.\egroup
  }{2018}]{Wachter2017}
Sandra Wachter, Brent Mittelstadt, and Chris Russell.
\newblock Counterfactual explanations without opening the black box: Automated
  decisions and the gdpr.
\newblock {\em Harvard Journal of Law and Technology}, 31(2):841--887, 2018.

\bibitem[\protect\citeauthoryear{Zhang \bgroup \em et al.\egroup
  }{2018}]{Zhang2018}
Xiangyu Zhang, Xinyu Zhou, Mengxiao Lin, and Jian Sun.
\newblock {ShuffleNet}: An extremely efficient convolutional neural network for
  mobile devices.
\newblock In {\em 2018 {IEEE}/{CVF} Conference on Computer Vision and Pattern
  Recognition}. {IEEE}, jun 2018.

\end{thebibliography}
